\documentclass[square,semicolon]{article} %
\usepackage[preprint]{neurips_2024}

\usepackage[utf8]{inputenc} %
\usepackage[T1]{fontenc}    %
\usepackage{hyperref}       %
\usepackage{url}            %
\usepackage{booktabs}       %
\usepackage{amsfonts}       %
\usepackage{nicefrac}       %
\usepackage{microtype}      %
\usepackage{xcolor}         %

\usepackage{graphicx}
\usepackage{subcaption}
\usepackage{amsmath}
\usepackage{adjustbox}
\usepackage{bera}%
\usepackage{listings}
\usepackage{xcolor}
\usepackage[frozencache,cachedir=minted-cache]{minted} 

\colorlet{punct}{red!60!black}
\definecolor{background}{HTML}{EEEEEE}
\definecolor{delim}{RGB}{20,105,176}
\colorlet{numb}{magenta!60!black}

\lstdefinelanguage{json}{
    basicstyle=\normalfont\ttfamily,
    numbers=left,
    numberstyle=\scriptsize,
    stepnumber=1,
    numbersep=8pt,
    showstringspaces=false,
    breaklines=true,
    backgroundcolor=\color{background},
    literate=
     *{0}{{{\color{numb}0}}}{1}
      {1}{{{\color{numb}1}}}{1}
      {2}{{{\color{numb}2}}}{1}
      {3}{{{\color{numb}3}}}{1}
      {4}{{{\color{numb}4}}}{1}
      {5}{{{\color{numb}5}}}{1}
      {6}{{{\color{numb}6}}}{1}
      {7}{{{\color{numb}7}}}{1}
      {8}{{{\color{numb}8}}}{1}
      {9}{{{\color{numb}9}}}{1}
      {:}{{{\color{punct}{:}}}}{1}
      {,}{{{\color{punct}{,}}}}{1}
      {\{}{{{\color{delim}{\{}}}}{1}
      {\}}{{{\color{delim}{\}}}}}{1}
      {[}{{{\color{delim}{[}}}}{1}
      {]}{{{\color{delim}{]}}}}{1},
}

\definecolor{darkblue}{rgb}{0, 0.2, 0.7}
\hypersetup{
    colorlinks = true,
    linkcolor = darkblue,
    anchorcolor = darkblue,
    citecolor = darkblue,
    filecolor = darkblue,
    urlcolor = darkblue
}

\usepackage{longtable}
\usepackage{array}
\renewcommand\arraystretch{1.5} %
\usepackage{listings}

\newcommand{\best}[1]{\textbf{\underline{#1}}}

\usepackage{wrapfig}    %

\usepackage{enumitem}

\usepackage{algorithm}
\usepackage{algpseudocode}

\newcommand{\denselist}{ \itemsep -2pt\topsep-10pt\partopsep-10pt }

\usepackage[most]{tcolorbox}
\tcbset{
  aibox/.style={
    width=\textwidth,
    top=10pt,
    colback=white,
    colframe=black,
    colbacktitle=black,
    enhanced,
    center,
    attach boxed title to top left={yshift=-0.1in,xshift=0.15in},
    boxed title style={boxrule=0pt,colframe=white,},
  }
}
\newtcolorbox{AIbox}[2][]{aibox,title=#2,#1}

\title{JSONSchemaBench: A Rigorous Benchmark of Structured Outputs for Language Models}

\author{%
  Saibo Geng$^{1}$\thanks{Work done during internship at Microsoft.} \quad Hudson Cooper$^{2}$ \quad Micha{\l} Moskal$^{2}$
  \quad Samuel Jenkins$^{2}$ \\
  \textbf{Julian Berman$^{3}$} 
 \quad \textbf{Nathan Ranchin$^{1}$} \quad \textbf{Robert West$^{1,2}$} \\
 \quad \textbf{Eric Horvitz$^{2}$} 
  \quad \textbf{Harsha Nori$^{2}$} \\
  $^1$EPFL \qquad $^2$Microsoft \qquad $^3$JSON Schema\\
  \{saibo.geng, nathan.ranchin, robert.west@epfl.ch\}@epfl.ch  \\
  \{julian\}@grayvines.com \\
  \{hanori, hudsoncooper, michal.moskal, sajenkin, horvitz\}@microsoft.com
}

\begin{document}

\newtheorem{definition}{Definition}[section]

\maketitle

\begin{abstract}
    Reliably generating structured outputs has become a critical capability for modern language model (LM) applications. {\em Constrained decoding} has emerged as the dominant technology across sectors for enforcing structured outputs during generation. 
    Despite its growing adoption, little has been done with the systematic evaluation of the behaviors and performance of constrained decoding.
    Constrained decoding frameworks have standardized around JSON Schema as a structured data format, with most uses guaranteeing constraint compliance given a schema. However, there is poor understanding of the effectiveness of the methods in practice.
    We present an evaluation framework to assess constrained decoding approaches across three critical dimensions: efficiency in generating constraint-compliant outputs, coverage of diverse constraint types, and quality of the generated outputs.
    To facilitate this evaluation, we introduce JSONSchemaBench, a benchmark for constrained decoding comprising 10K real-world JSON schemas that encompass a wide range of constraints with varying complexity. 
    We pair the benchmark with the existing official JSON Schema Test Suite and evaluate six state-of-the-art constrained decoding frameworks, including \textit{Guidance}, \textit{Outlines}, \textit{Llamacpp}, \textit{XGrammar}, \textit{OpenAI}, and \textit{Gemini}.
    Through extensive experiments, we find that JSONSchemaBench presents a significant challenge for both LLMs and constrained decoding frameworks, highlighting ample room for improvement and exposing gaps in the existing solutions.
    We release JSONSchemaBench at \url{https://github.com/guidance-ai/jsonschemabench}.
\end{abstract}

\section{Introduction}\label{sec:introduction}

The rapid advancements in LMs in recent years have significantly broadened their applications, extending beyond natural language tasks to more complex challenges such as web navigation~\citep{yao2023reactsynergizingreasoningacting}, data extraction~\citep{Polak_2024}, and tool use~\citep{schick_toolformer_2023}.
Unlike traditional natural language processing (NLP) tasks where the output is aimed at review by humans, output in these applications is often consumed by machines such as controller and service APIs.
The machine-oriented nature of these applications requires LMs to generate structured outputs that strictly adhere to predefined formats and constraints.
However, the LM generation process is probabilistic and does not provide guarantees on the output's structure, making it challenging to deploy LMs in applications requiring structured inputs and high reliability.

\begin{figure}[t]
  \centering
  \includegraphics[width=0.7\textwidth]{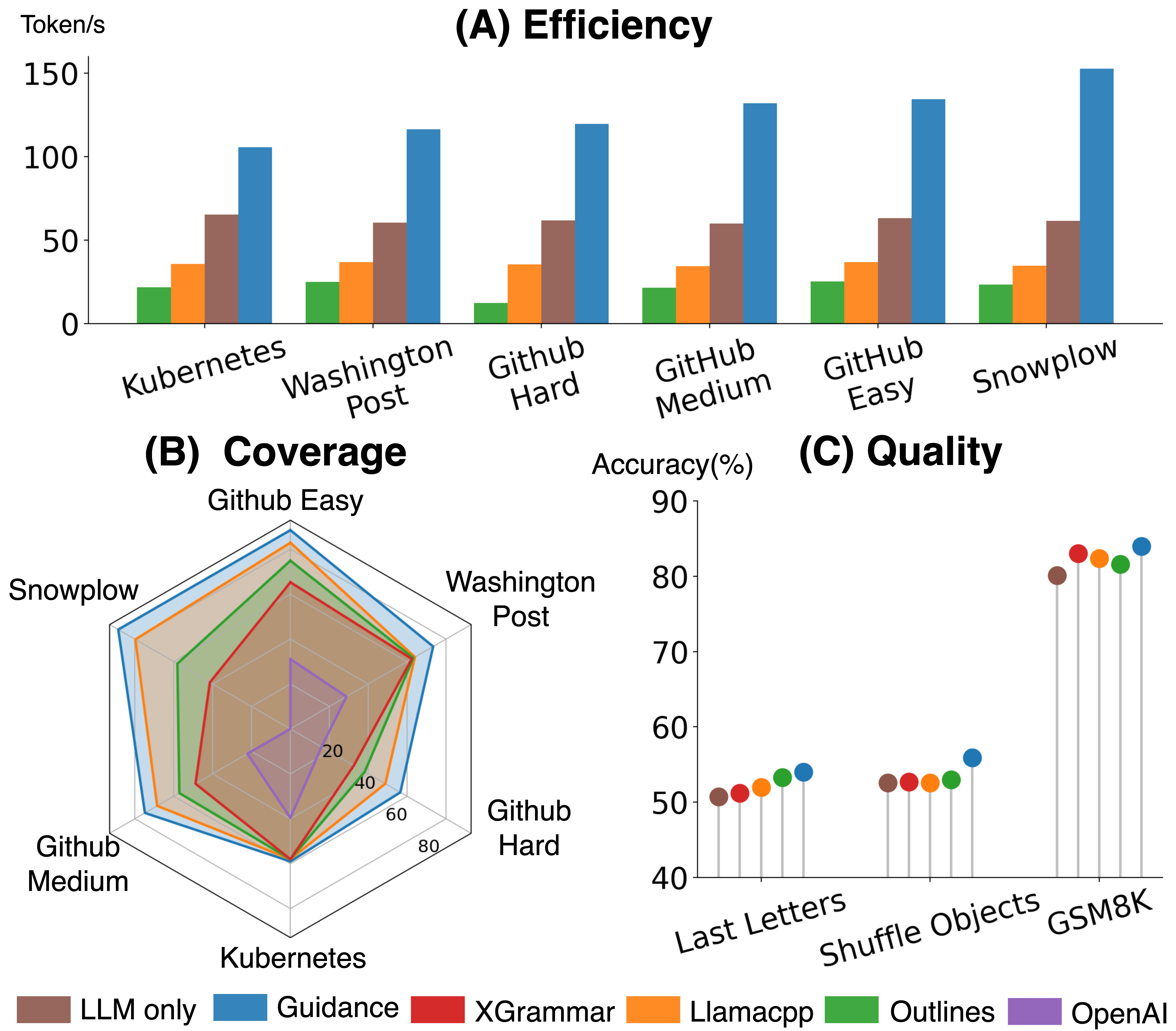}
  \caption{Comparison across various constrained-decoding frameworks by
efficiency (speed of output generation), coverage (support for JSON Schema features), and quality (effects on underlying task accuracy). Guidance outperforms other frameworks on these dimensions.}
  \label{fig:overview}
\end{figure}

The methodology of constrained decoding, a technique that integrates constraints into the decoding process of LMs, has been developed to address the need to adapt LM generations to the challenge of providing structured output.
Constrained decoding intervenes in the decoding process of LMs by masking out invalid tokens based on given constraints and prefix tokens. 
This intervention guides the LM to sample only from valid tokens, ensuring that the final output perfectly conforms to a predefined structure.

The strong demand for structured generation~\citep{liu_we_2024} has led to the development of various constrained-decoding frameworks\footnote{We use the terms \emph{constrained decoding framework} and  \emph{grammar engine} interchangeably.}, such as Guidance~\citep{guidance}, Outlines~\citep{willard_efficient_2023}, XGrammar~\citep{dong_xgrammar_2024} and the grammar module of Llamacpp~\citep{llama_cpp}
These frameworks provide broad support for different types of constraints, minimal overhead, and compatibility with various LM ecosystems, facilitating the adoption of constrained decoding in real-world applications.

JSON Schema offers a high level, domain-specific way to define constraints for JSON data, a widely adopted data interchange format. As a result, JSON Schema has emerged as a key specification language for constrained decoding.
Commercial LM providers, such as OpenAI, have embraced constrained decoding by incorporating support for JSON Schema directly into their APIs. These integrations highlight the emergence of JSON Schema as an industry-wide standard for specifying constraints on structured outputs, ensuring compatibility across diverse applications.

Despite the growing adoption of constrained decoding for structured generation, several issues and questions persist:\\
\textbf{Q1: Efficiency}: Does constrained decoding slow down or speed up the generation process? Which framework is the most efficient?\\
\textbf{Q2: Coverage}: The JSON Schema specification has an evolving and expansive feature set. How well do existing constrained decoding frameworks support these features? \\
\textbf{Q3: Quality}: While constrained decoding guarantees that LM outputs conform to a desired structure, does it negatively affect the semantic quality of outputs?

To answer these questions, we need to study constrained-decoding methods with a large-scale, diverse, and real-world collection of user-defined structures. To evaluate the performance of constrained decoding frameworks, we introduce \textbf{JSONSchemaBench}, a collection of 10K real-world JSON schemas from various sources, 
Organized into 10 datasets of varying complexity and diversity, the benchmark spans domains such as function signatures, service APIs, and system configurations.
We evaluate six state-of-the-art constrained decoding frameworks, including Guidance, Outlines, Llamacpp, XGrammar, OpenAI, and Gemini, on JSONSchemaBench.
We pair this real-world schema dataset with the official JSON Schema Test Suite~\citep{jsonSchemaTestSuite} in order to extract detailed insights into coverage of JSON Schema functionality across these frameworks, and to further evaluate them with considerations of end-to-end task accuracy in the context of multiple real-world tasks.
Altogether, our evaluation takes three aspects into consideration: efficiency, coverage,  and quality. We define specific metrics to measure these three functional aspects and evaluate constrained decoding frameworks against them.
Through extensive experiments, we converge on the following findings as illustrated in Figure~\ref{fig:overview}.
(1) Constrained decoding can speed up the generation process by 50\% compared to unconstrained decoding.
(2) Frameworks demonstrate significant differences in their actual support for real-world JSON schemas, with the best framework supporting twice as many schemas as the worst.
(3) Constrained decoding consistently improves the performance of downstream tasks up to 4\%, even for tasks with minimal structure like GSM8k.

\paragraph{Contributions}
Our contributions are three-fold:
\begin{itemize}
\denselist
    \item We assemble JSON schemas from various sources and organize them into a benchmark, JSONSchemaBench, to facilitate the evaluation of constrained decoding frameworks on JSON schema.
    \item We propose a fine-grained evaluation framework to assess the versatility of constrained decoding frameworks in handling diverse JSON schema features, including declared coverage, empirical coverage, and compliance rate.
    \item We evaluate six state-of-the-art constrained decoding frameworks on JSONSchemaBench, uncovering their strengths and limitations in generating schema-compliant JSON outputs and analyzing their impact on downstream tasks.
\end{itemize}

\section{Background and Related Work}\label{sec:background_related_work}

\emph{JSON Schema} is a meta-language that describes the structure of JSON data.
It is capable of expressing a wide variety of constraints, such as the types of JSON object properties, the length of JSON arrays or the pattern that a JSON string must match.
The syntax and capabilities of JSON Schema are defined in the JSON Schema specification~\citep{jsonschema2020draft}, which defines a large number of \textit{keywords}, each of which may be used or combined with other keywords within a schema to enforce constraints like the ones mentioned. JSON Schema is widely used in the software ecosystem, and previous work has been done to collect extensive examples of JSON Schemas with a focus both on real-world use as well as on overall correctness.

\cite{schema_corpus} collected over 6,000 JSON schemas from publicly available GitHub repositories. \cite{attouche_witness_2022} used it alongside additional collected JSON schemas in order to evaluate a witness generation algorithm for JSON Schema. Separately, the official JSON Schema Test Suite~\citep{jsonSchemaTestSuite} is a collection of manually created test cases, maintained by the JSON Schema core team, which exercises a large portion of the functionality defined in the JSON Schema specification. It was originally written to assist implementers of JSON Schema validation tools with testing their compliance against the specification, and therefore contains a wide variety of examples for each of JSON Schema's keywords, including in edge case scenarios. Notably, Bowtie~\citep{bowtie} leverages the test suite as a foundation for comparing and understanding different implementations of the JSON Schema specification across programming languages. Taken together, these two datasets form a large number of examples both of JSON Schema's diverse feature set as well as its use in the wild.

\begin{wrapfigure}{R}{0.5\textwidth} %
    \begin{minipage}{0.5\textwidth} %
        \begin{algorithm}[H]
            \caption{Constrained Decoding}
            \label{alg:constrained_decoding}
            \begin{algorithmic}[1]
                \Require Constraint $C$, LLM $f$, Prompt $x$
                \Ensure Output $o$ adhering to $C$
                \State $o \gets []$
                \Loop
                    \State $C.\texttt{update}(o)$ \Comment{advance state of $C$}
                    \State $m \gets C.\texttt{mask}()$ \Comment{compute mask}
                    \State $v \gets f(x + o)$ \Comment{compute logits}
                    \State $v' \gets m \odot v'$ 
                    \State $t \gets \texttt{decode}(\alpha')$ \Comment{sample}
                    \If{$t = \text{EOS}$}
                        \State \textbf{break}
                    \EndIf
                    \State $o.\texttt{append}(t)$
                \EndLoop
                \State \Return $o$ \Comment{ output}
            \end{algorithmic}
        \end{algorithm}
    \end{minipage}
\end{wrapfigure}

Constrained decoding~\citep{deutsch_general-purpose_2019,shin_constrained_2021, scholak_picard_2021, poesia_synchromesh_2022,wang_grammar_nodate, geng-etal-2023-grammar} refers to methods that guide the generation process of language models (LMs) by masking out tokens that do not adhere to predefined constraints at each step. 
Recently, highly optimized grammar-constrained decoding frameworks~\citep{guidance,beurer-kellner_prompting_2023,willard_efficient_2023, kuchnik_validating_2023,zheng_sglang_2024,dong_xgrammar_2024} have been developed to improve the efficiency and usability of constrained decoding.

The evaluation of constrained decoding remains an under-explored topic, with no consensus on what defines the effectiveness of constrained decoding. While some research has pursued comparisons of constrained decoding with unconstrained LMs~\citep{roy_benchclamp_2024,tang_struc-bench_2024,yao_collie_2023}, the studies to date fail to provide comparisons across different constrained decoding frameworks. The benchmarks employed have either narrowly focused on specific tasks or rely on formal-grammar--based artificial setups, that have unclear relevance to real-world use cases.

\section{The JSONSchemaBench}\label{sec:benchmark}

Our goal is to design a benchmark that is
(1) diverse enough to cover the most common constraint types encountered in real-world applications, (2) large enough to provide a reliable evaluation, and (3) equipped with fair and multidimensional metrics to ensure comprehensive assessments.

\subsection{Data Collection}

We start with the 6K JSON schemas collected by \cite{schema_corpus} from publicly available GitHub repositories, and with the set of schemas from the JSON Schema Test Suite~\citep{jsonSchemaTestSuite}.
We further collect JSON schemas from other sources, such as the JSON Schema Store~\citep{jsonschemastore}, the GlaiveAI function calling dataset V2~\citep{glaiveai2024functioncalling}, and from Kubernetes configuration files~\citep{kubernetes2022schemas}.
We filter out invalid schemas and standardize the schemas to ensure that they conform to the version of JSON Schema declared\footnote{The \texttt{\$schema} keyword, defined in the JSON Schema specification, allows any schema to self-identify which version of JSON Schema it is written for.} in each schema
\begin{wraptable}{r}{0.5\textwidth} %
    \centering
    \renewcommand{\arraystretch}{0.9}
    \setlength{\tabcolsep}{4pt} %
    \renewcommand{\arraystretch}{1.2} %
    {\scriptsize %
    \caption{Schema collection metadata.}
    \begin{tabular}{@{}lll@{}}
        \toprule
        \textbf{Dataset} & \textbf{Category} & \textbf{Count} \\ 
        \midrule
        GlaiveAI-2K       & Function Call    & 1707  \\
        Github-Trivial    & Misc             & 444  \\      
        Github-Easy       & Misc             & 1943  \\
        Snowplow          & Operational API       & 403   \\
        Github-Medium     & Misc             & 1976  \\
        Kubernetes        & Kubernetes API     & 1064  \\
        Washington Post   & Resource Access API      & 125   \\
        Github-Hard       & Misc             & 1240  \\
        JSONSchemaStore   & Misc             & 492   \\
        Github-Ultra      & Misc             & 164   \\
        \midrule
        \textbf{Total}     &                  & 9558  \\
        \bottomrule
    \end{tabular}
    }
    \label{tab:dataset_description_short}
\end{wraptable}

The GitHub JSON schemas collection from \cite{schema_corpus} contains schemas of varying complexity and diversity, ranging from simple type constraints to complex constraints with nested objects and arrays.
For more fine-grained evaluation, we split the data into five collections based on the schema size: trivial, small, medium, large, ultra.
The suites finalized after all collection and processing are listed in Table~\ref{tab:dataset_description_short}.
We excluded GitHub-Trivial and GitHub-Ultra from the experiments as they were too easy or too hard. However, we retained these datasets in the benchmark, with GitHub-Ultra serving as an aspirational target for future advancements.
For more information on post-processing and dataset splitting, we refer the reader to Appendix~\ref{app:schema_collection_details}.

\section{Efficiency}

Naïve implementations of constrained decoding add overhead to the standard LM inference process, including a per-step mask computation and an optional one-time grammar compilation.
However, several optimizations can significantly reduce this overhead. For instance, mask computation can run in parallel with the LM's forward pass, and grammar compilation can be performed concurrently with pre-filling computations~\citep{guidance, dong_xgrammar_2024}.
Other optimizations such as grammar caching and constraint-based speculative decoding~\citep{guidance_acceleration, beurer-kellner_prompting_2023,coalescence2024} can further reduce overhead.

\paragraph{Metrics}
We break down the efficiency evaluation into the following components:
\begin{itemize}
\denselist
    \item \textbf{Grammar Compilation Time (GCT):} The time spent on grammar compilation, if applicable.
    \item \textbf{Time to First Token (TTFT):} Time from the start of generation to the production of the first token.
    \item \textbf{Time per Output Token (TPOT):} Average time to generate each output token after the first.
\end{itemize}

\subsection{Setup}

The efficiency experiment depends on both the size of the model and the tokenizer's vocabulary size. 
We used \textbf{Llama-3.1-8B-Instruct} with the \textbf{Llamacpp} inference engine as backend for Outlines, Guidance, and Llamacpp.
As XGrammar doesn't support Llamacpp as backend , we add an additional experiment with the \textbf{Hugging Face Transformers} inference engine for XGrammar.
All experiments are conducted on a single \textbf{NVIDIA A100-SXM4-80GB GPU} with \textbf{AMD EPYC 7543 (12 cores)} CPU.
The batch size is set to 1 for all experiments.
Additional details about setup are provided in the Appendix~\ref{app:efficiency_experiment_details}.
We also provide a snippet of how we call each engine in the Appendix~\ref{app:engine_snippet}.

\paragraph{Addressing coverage bias.} The efficiency metrics are meaningful only for instances that a grammar engine can process. Different engines exhibit varying levels of schema coverage, with some engines handling a wider range of schemas than others. Engines with lower coverage often process simpler, shorter schemas, which naturally compile and generate faster. As a result, averaging efficiency metrics across covered instances can introduce bias favoring engines with lower coverage. For a more detailed discussion on coverage, see Section~\ref{sec:coverage}.
To ensure fairness, we calculate efficiency metrics on the intersection of covered instances across all engines.

\begin{table}[h!]
    \centering
    \small
    \caption{\textbf{Efficiency metrics} for different engines with \textbf{LlamaCpp} as the inference engine. 
    \textbf{GCT}: Grammar Compilation Time, \textbf{TTFT}: Time to First Token, \textbf{TPOT}: Time Per Output Token.
    Bold values indicate the smallest in each column for GCT, TTFT, and TPOT. All values are \textbf{median} of the samples. Results for the GitHub Hard and Washington Post datasets are provided in Appendix~\ref{app:efficiency_experiment_details}.}
    \label{tab:efficiency}
    \renewcommand{\arraystretch}{0.9}
    \begin{adjustbox}{max width=\textwidth}
    \begin{tabular}{llrrr}
    \toprule
    \textbf{Dataset} & \textbf{Framework} & \textbf{GCT (s)} & \textbf{TTFT (s)} & \textbf{TPOT (ms)} \\
    \midrule
    \textbf{GlaiveAI}        & LM only     &    NA         & \best{0.10}   &       15.40\\
                             & Guidance     & \best{0.00}   & 0.24          & \best{6.37}\\
                             & Llamacpp     & 0.05          & 0.20          & 29.98\\
                             & Outlines     & 3.48          & 3.65          & 30.33\\
    \cmidrule(lr){1-5}
    \textbf{GitHub Easy}     & LM only     & NA            & \best{0.10}   &       15.83\\
                             & Guidance     & \best{0.00}   & 0.34          & \best{7.44}\\
                             & Llamacpp     & 0.05          & 0.18          & 27.22\\
                             & Outlines     & 3.71          & 3.97          & 39.78\\
    \cmidrule(lr){1-5}
    \textbf{Snowplow}        & LM only     &    NA         & \best{0.11}   &       16.23\\
                             & Guidance     & \best{0.00}   & 0.28          & \best{6.55}\\
                             & Llamacpp     & 0.05          & 0.20          & 28.90\\
                             & Outlines     & 3.91          & 4.14          & 42.66\\
    \cmidrule(lr){1-5}
    \textbf{GitHub Medium}   & LM only     &    NA         & \best{0.20}   &       16.68\\
                             & Guidance     & \best{0.01}   & 0.54          & \best{7.57}\\
                             & Llamacpp     & 0.06          & 0.30          & 29.08\\
                             & Outlines     & 8.05          & 8.38          & 46.57\\
    \cmidrule(lr){1-5}
    \textbf{Kubernetes}      & LM only     &    NA         &  \best{0.16}  &       15.32\\
                             & Guidance     & \best{0.01}   & 0.45          & \best{9.47}\\
                             & Llamacpp     & 0.05          & 0.28          & 28.04\\
                             & Outlines     & 5.29          & 5.55          & 46.10 \\
    \bottomrule
    \end{tabular}
    \end{adjustbox}
\end{table}

\paragraph{Grammar compilation time.}

There are notable differences in grammar compilation times between the engines. 
Both Guidance and Llamacpp dynamically compute their constraints during token generation, leading to minimal grammar compilation time. In the middle, XGrammar does include a non-trivial compilation step, but they are able to largely mitigate its impact by running it concurrently with prompt pre-filling. Finally Outlines, which converts JSON schemas into regular-expression based constraints, has significantly higher compilation time.

\paragraph{Time per output token.}
While Outlines and Llamacpp demonstrate substantially lower throughput than the LM-only approach, Guidance achieves even higher efficiency, which it accomplishes by fast-forwarding\footnote{See Tables\ref{tab:efficiency_llamacpp_additional} and~\ref{tab:efficiency_hf_additional} for the number of tokens fast-forwarded in the experiments.} certain generation steps with its \textit{guidance acceleration} ~\citep{guidance_acceleration}. Comparing Guidance and XGrammar with the HF Transformers backend shows that Guidance has a significantly better TPOT. 

\begin{table}[h!]
    \centering
    \caption{As XGrammar doesn’t support
    \textbf{llama.cpp}, we add an additional experiment with the \textbf{Hugging Face Transformers}
    inference engine for XGrammar and Guidance. All values are \textbf{median} of the result samples. }
    \label{tab:efficiency_hf}
    \renewcommand{\arraystretch}{0.9}
    \begin{adjustbox}{max width=\textwidth}
    \begin{tabular}{llrrr}
    \toprule
    \textbf{Dataset} & \textbf{Framework} & \textbf{GCT (s)} & \textbf{TTFT (s)} & \textbf{TPOT (ms)} \\
    \midrule
    \textbf{GlaiveAI}      
                            & Guidance      & \best{0.01}   & 0.36          & \best{36.92}\\
                            & XGrammar      & 0.12          & \best{0.30}   & 66.78       \\
                            
    \cmidrule(lr){1-5}
    \textbf{GitHub Easy}   
                            & Guidance      & \best{0.01}   & 0.37          & \best{42.03}\\
                            & XGrammar      & 0.11          & \best{0.33}   & 65.57       \\
                            
    \cmidrule(lr){1-5}
    \textbf{GitHub Medium}  
                            & Guidance      & \best{0.01}   & 0.55          & \best{44.21}\\
                            & XGrammar      & 0.20          & \best{0.48}   & 65.51       \\
    \cmidrule(lr){1-5}
    \textbf{GitHub Hard}    
                            & Guidance      & \best{0.01}   & 0.73          & \best{35.88}\\
                            & XGrammar      & 0.30          & \best{0.65}   & 65.20       \\
    \bottomrule
    \end{tabular}
    \end{adjustbox}
    \end{table}

\section{Coverage}\label{sec:coverage}

Each constrained decoding framework has limitations when it comes to translating JSON schemas into a set of constraints that can reliably guarantee the validity of LM outputs. To systematically evaluate the effectiveness of these frameworks, we define three notions of coverage:

\begin{definition}[Declared Coverage]
    A schema is considered \textit{declared covered} if the framework processes the schema without explicitly rejecting it or encountering runtime errors such as exceptions or crashes.
\end{definition}

\begin{definition}[Empirical Coverage]
    A schema is considered \textit{empirically covered} if our experiments show that the constraints generated by the framework result in LM outputs that are schema-compliant. 
\end{definition}

\begin{definition}[True Coverage]
    A schema is considered \textit{truly covered} if the framework produces constraints that are precisely equivalent to the original JSON Schema definition, i.e., permitting all schema-compliant generations while rejecting all schema-noncompliant generations.
\end{definition}

The most ideal coverage metric is the \textit{true coverage}, denoted as $\cal{C}_{\text{True}}$. However, due to the infinite number of JSON instances that could be validated against a schema, it is difficult to measure in practice without a formal verification method that is capable of exhaustively comparing the schema's semantics against the framework's implementation.
$\cal{C}_{\text{Empirical}}$ is an approximation of $\cal{C}_{\text{True}}$ as it only checks whether the finitely many outputs seen during our experiments conform to a given schema\footnote{Additionally, we define \textit{theoretical coverage} as the proportion of schemas whose features are fully supported by the grammar engine, with details provided in Appendix~\ref{app:theoretical_coverage_details}.}.

While $\cal{C}_{\text{Declared}}$ is not an estimate of $\cal{C}_{\text{True}}$ per se, it is an upper-bound of both $\cal{C}_{\text{Empirical}}$ and $\cal{C}_{\text{True}}$ and is useful in deriving an additional metric from the coverage evaluation: \textbf{Compliance Rate} $= \cal{C}_{\text{Empirical}} / \cal{C}_{\text{Declared}}$.
The \textit{Compliance Rate} estimates the reliability of the constrained decoding framework in guaranteeing compliance given it accepts a given schema.

\subsection{Setup}

To measure empirical coverage, we conduct all experiments using the Llama-3.2-1B-Instruct model as it is small enough to run efficiently while still producing high-quality outputs.
The prompt consists of a simple instruction with two-shot examples (Figure~\ref{fig:json-prompt-template}), and validation is performed using the \texttt{jsonschema} Python library (\cite{python-jsonschema}) (using JSON Schema Draft2020-12) with string-format checks enabled. 
We use greedy decoding with zero-temperature, performing a single generation run, and enforce a 40-second timeout for grammar compilation and an additional 40 seconds for generation. 
Exceeding these limits is treated as a schema processing failure. 
Additional details are provided in Appendix~\ref{app:coverage_experiment_details}.

\subsection{Results}\label{subsec:coverage_results}

\paragraph{Empirical Coverage}
Guidance shows the highest empirical coverage on six out of the eight datasets, with Llamacpp taking the lead on the remaining two: the domain-specific Washington Post and notably hard JSON Schema Store.
On the other hand, closed-source grammar engines consistently have the lowest coverage; they came in last on all but one dataset.
LM-only\footnote{The Llama 3.1 models have been specifically fine-tuned to adhere to JSON schemas~\citep{grattafiori2024llama3herdmodels}} approaches achieve acceptable coverage on easy-to-medium datasets but show significant performance drops on harder datasets, such as Github Hard and JSON Schema Store, as well as domain-specific datasets like Washington Post.
We note that while empirical coverage is a reasonable indicator of a framework's real-world performance, it is influenced by factors such as the LM being used and the sampling methods employed.

\renewcommand{\arraystretch}{0.85}
\begin{table}[h!]
    \centering
    \caption{\textbf{Coverage of all the frameworks} on JSONSchemaBench. Empirical coverage between Open Source engines and OpenAI/Gemini are not directly comparable due to differences in the underlying model (Llama 3.2-1B vs. proprietary models). \\
    \footnotesize{$^*$ Gemini results are ommitted for dataset suites with $< 1\%$ support.}}
   \label{tab:coverage}
    \begin{adjustbox}{max width=\textwidth}
    {\footnotesize
    \begin{tabular}{llrrr}
    \toprule
    \textbf{Dataset} & \textbf{Framework}& \textbf{Declared} & \textbf{Empirical} & \textbf{Compliance} \textbf{Rate}  \\
    \midrule
    \textbf{GlaiveAI}        &  LM only & 1.00  & 0.90  & 0.90  \\
                             &  Guidance & 0.98 & \best{0.96} & \best{0.98} \\
                             &  Llamacpp & 0.98 & 0.95 & 0.97 \\
                             &  Outlines & 0.99 & 0.95 & 0.96 \\
                             &  XGrammar & 1.00  & 0.93  & 0.93   \\
   \cmidrule(lr){2-5}
                             &  OpenAI & 0.89   & 0.89  & 1.00      \\
                             &  Gemini & 0.86   & 0.86  & 1.00   \\
    \cmidrule(lr){1-5}
    \textbf{GitHub Easy}    &  LM only & 1.00  & 0.65  & 0.65  \\
                             &  Guidance & 0.90 & \best{0.86} & \best{0.96} \\
                             &  Llamacpp & 0.85 & 0.75 & 0.88 \\
                             &  Outlines & 0.86 & 0.59 & 0.83 \\
                             &  XGrammar & 0.91  & 0.79  & 0.87   \\
   \cmidrule(lr){2-5}
                            &  OpenAI & 0.30   & 0.29  & 0.97   \\
                            &  Gemini & 0.08   & 0.07  & 0.88   \\
    \cmidrule(lr){1-5}
    \textbf{Snowplow$^*$}       &  LM only & 1.00  & 0.46  & 0.46  \\
                             &  Guidance & 0.87 & \best{0.82} & \best{0.94}  \\
                             &  Llamacpp & 0.92 & 0.74 & 0.81  \\
                             &  Outlines & 0.95 & 0.36 & 0.61  \\
                                &  XGrammar & NA  & NA  & NA    \\
   \cmidrule(lr){2-5}
                                &  OpenAI & 0.21   &  0.21 & 1.00    \\
    \cmidrule(lr){1-5}
    \textbf{GitHub Medium$^*$}   &  LM only & 1.00  & 0.38  & 0.38  \\
                             &  Guidance & 0.79 & \best{0.69} & \best{0.87} \\ 
                             &  Llamacpp & 0.77 & 0.57 & 0.74  \\
                             &  Outlines & 0.72 & 0.29 & 0.40  \\
                                &  XGrammar & 0.79  & 0.52  & 0.66    \\
   \cmidrule(lr){2-5}
                                &  OpenAI & 0.13  & 0.12  & 0.92    \\
    \cmidrule(lr){1-5}
    \textbf{Kubernetes$^*$}      &  LM only & 1.00  & 0.56 &      0.56 \\
                             &  Guidance & 0.98 & \best{0.91} & \best{0.92}  \\
                             &  Llamacpp & 0.98 & 0.76 & 0.78  \\
                             &  Outlines & 0.98 & 0.57 & 0.58  \\
                                &  XGrammar & 0.12  & 0.07  & 0.58    \\
   \cmidrule(lr){2-5}
                                &  OpenAI & 0.21   & 0.21   & 1.00      \\
    \cmidrule(lr){1-5}
    \textbf{Washington Post$^*$} &  LM only & 1.00  & 0.40  & 0.40  \\
                             &  Guidance & 0.86 & 0.86 & \best{1.00}  \\
                             &  Llamacpp & 0.97 & \best{0.94} & 0.97  \\
                             &  Outlines & 0.97 & 0.22 & 0.23  \\
                                &  XGrammar & 0.85  & 0.64  & 0.75    \\
   \cmidrule(lr){2-5}
                                &  OpenAI & 0.13    & 0.13  & 1.00    \\
    \cmidrule(lr){1-5}
    \textbf{GitHub Hard$^*$}    &  LM only & 1.00  & 0.13  & 0.13  \\
                            &  Guidance & 0.60 & \best{0.41} & \best{0.69}  \\
                             &  Llamacpp & 0.61 & 0.39 & 0.63  \\
                             &  Outlines & 0.47 & 0.03 & 0.06  \\
                                &  XGrammar & 0.69  & 0.28  & 0.41    \\
   \cmidrule(lr){2-5}
                                &  OpenAI & 0.09    & 0.09  & 1.00    \\
    \cmidrule(lr){1-5}
    \textbf{JsonSchemaStore$^*$} &  LM only & 1.00  & 0.21  & 0.21  \\
                             &  Guidance & 0.35 & 0.30 & \best{0.88}  \\
                             &  Llamacpp & 0.54 & \best{0.38} & 0.69  \\
                             &  Outlines & 0.38 & 0.09 & 0.24  \\
                                &  XGrammar & 0.76  & 0.33  & 0.43    \\
       \cmidrule(lr){2-5}
                                &  OpenAI & 0.06    & 0.06  & 1.00    \\
    \bottomrule
    \end{tabular}
    }
    \end{adjustbox}
\end{table}
\renewcommand{\arraystretch}{1.00}

\paragraph{Compliance Rate}
Among open-source engines, guidance consistently demonstrates the highest compliance rate across all datasets, making it the most reliable option for ensuring schema compliance.
Outlines has a comparatively lower compliance rate, primarily due to timeouts during generation. Our analysis reveals that JSON Schema features like `minItems`, `maxItems`, `enum`, and `Array', while supported, often take 40 seconds to 10 minutes for Outlines to process.
LM-only exhibits the lowest compliance rate, highlighting its unreliability as a standalone solution.
While closed-source implementations have low empirical coverage, they have very high compliance rates, indicating that their providers have taken a more conservative strategy, implementing only a subset of JSON Schema features that they can reliably support.

\subsection{JSON Schema Test Suite: Complementary Evaluation}\label{subsec:testsuite}

Originally designed to test the correctness and compliance of JSON Schema validation implementations, the official JSON Schema Test Suite~\citep{jsonSchemaTestSuite} is a comprehensive collection of test cases spanning the many features of the JSON Schema specification. We believe that the test suite is an ideal tool for assessing the correctness of grammar engines.

The test suite organizes its test cases into 45 categories, each of which corresponds to a feature of JSON Schema, typically a specific keyword such as \texttt{required} or group of tightly related keywords such as \texttt{if-then-else}. A small number of additional categories test broader behaviors, such as \texttt{infinite-loop-detection}.
Each test case contains a single schema paired with a collection of JSON instances that are marked as either valid or invalid under that schema. For the purpose of evaluating coverage, we assert that an engine must successfully generate each valid instance and block generation of each invalid instance to ``pass'' a test case. In addition to compilation failures, we define two failure modes that a grammar engine can exhibit:

\begin{definition}[Over-constrained]
    A framework is \emph{over-constrained} if it rejects JSON instances that are valid according to a given JSON Schema. This means the engine is too strict and excludes outputs that should be allowed.
\end{definition}
\begin{definition}[Under-constrained]
    A framework is \emph{under-constrained} if it allows JSON instances that are invalid according to a given JSON Schema. This means the engine is overly permissive and allows outputs that should be rejected.
\end{definition}
An illustration is given in Figure~\ref{fig:venn_diagram_over_under} in Appendix~\ref{app:test_suite_experiment_details}.
\textit{Over-constrained} grammar engines risk limiting the expressive power of LMs, potentially preventing the generation of valid responses and negatively impacting downstream task performance. 
Conversely, under-constrained engines cannot guarantee that all responses will be valid, often necessitating additional post-processing or retry logic.

\subsubsection{Results}

\paragraph{Coverage Analysis}
For each grammar engine and category in the test suite, we calculate \textit{test coverage} as the proportion of passing test cases, reported in Figure~\ref{fig:test_coverage} in Appendix~\ref{app:test_suite_experiment_details}
Additionally, Table~\ref{tab:test_coverage_agg} aggregates these metrics, counting categories with minimal coverage (\ensuremath{>} 0\%), partial coverage (\ensuremath{>} 25\%), moderate coverage (\ensuremath{>} 50\%), high coverage (\ensuremath{>} 75\%), and full coverage (100\%). We indicate the number of categories for which each framework achieves the highest test coverage (either as the single highest or as the sole leader) as well as the number of categories for which each framework is the sole leader.

\begin{itemize}
\denselist
    \item \textbf{Overall Performance}: Guidance outperforms other engines at all coverage levels, achieving full coverage on 13 categories and moderate coverage on 21. In comparison, Llamacpp and XGrammar have full coverage on only one category and moderate coverage on five and three categories, respectively, while Outlines has no full coverage on any category and moderate coverage on two categories. 
    \item \textbf{Single Highest}: Guidance has the single highest coverage in 19 categories, followed by XGrammar with 10, and Outlines with one, and Llamacpp with none.
\end{itemize}

\begin{table}[h!]
    \centering
    \caption{Number of categories with a given level of coverage. Each row represents a cumulative coverage threshold, with higher thresholds indicating stricter levels of success. Bold numbers indicate the framework with the highest value in that row.}
    \label{tab:test_coverage_agg}
    \renewcommand{\arraystretch}{0.9}
    \begin{adjustbox}{max width=\textwidth}
        \begin{tabular}{lcccc}
            \toprule
            Coverage & Outlines & Llamacpp & XGrammar & Guidance \\
            \midrule
            Minimal coverage (\ensuremath{>}0\%)    & 20    & 21    & 28    & \best{30} \\
            Partial coverage (\ensuremath{>}25\%)   & 11    & 11    & 16    & \best{25} \\
            Moderate coverage (\ensuremath{>}50\%)  & 2     & 5     & 3     & \best{21} \\
            High coverage (\ensuremath{>}75\%)      & 0     & 2     & 1     & \best{17} \\
            Full coverage (100\%)                   & 0     & 1     & 1     & \best{13} \\
            \midrule
            Tied for highest (\ensuremath{>}0\%)    & 4     & 6     & 14    & \best{25} \\
            Single highest                          & 1     & 0     & 10    & \best{19} \\
            \bottomrule
        \end{tabular}
    \end{adjustbox}
\end{table}

\paragraph{Failure Analysis}

Table~\ref{tab:test_failures_agg} provides a breakdown of failure modes for each framework across the test suite, detailing the number of categories with compilation errors, failures to generate positive instances (over-constrained), and failures to block negative instances (under-constrained).
\begin{table}[H]
    \centering
    \caption{Number of categories for which each failure type occurred at least once. Columns do not necessarily sum to the total number of categories, as some categories may have more than one failure type or no failures at all. Bold numbers indicate the framework with the fewest number of failures of a given type.}
    \label{tab:test_failures_agg}
    \begin{adjustbox}{max width=\textwidth}
        \begin{tabular}{lcccc}
            \toprule
            Failure type & Outlines & Llamacpp & XGrammar & Guidance \\
            \midrule
            Compile Error       & 42    & 37    & \best{3}     & 25 \\
            Over-constrained     & 16    & 18    & \best{5}     & 7 \\
            Under-constrained    & 8     & 7     & 38    & \best{1} \\
            \bottomrule
        \end{tabular}
    \end{adjustbox}
\end{table}

Overall, Guidance demonstrates the fewest total failures, in particular minimizing under-constrained errors.
Outlines, Llamacpp, and Guidance follow a consistent failure pattern, with most errors occurring during compilation and over-constrained failures being more frequent than under-constrained ones. In contrast, XGrammar minimizes compilation errors but shows the highest number of under-constrained failures, indicating a trade-off favoring permissiveness.

We acknowledge that there is no straightforward correspondence between test suite performance and empirical coverage. One reason for this is that not all features are equally represented in real-world schemas. As a result, strong or weak performance on specific features can have disproportionate impacts depending on their prevalence. Another reason is under-constraining effectively delegates responsibility to the LM, which may produce valid output despite a lack of strict constraints. We emphasize that while under-constraining can be a legitimate strategy, it requires careful implementation and transparency to ensure reliability.

\section{Quality}

In principle, constrained decoding should not affect the quality of the generated output as it only filters out the invalid tokens. 
However, things become more complicated due to ambiguity of tokenization~\citep{vivien2024regexconstraints, guidance2024tokenhealing, geng2024bytebpetokenizationinverse} and the distributional shifts caused by the intervention~\citep{geng-etal-2023-grammar,tam2024letspeakfreelystudy}.
As a hypothetical toy example, an LM might answer `89,000' instead of the correct `89000' in a GSM8K question. 
Constrained decoding can block the invalid token `,', enforcing structural compliance but potentially may cause the LM to go out of distribution and generate `890000' instead.
\cite{kurt2023say} argued that the performance decline observed in previous studies~\citep{tam2024letspeakfreelystudy} comes from inadequate prompting, insufficient contextual information, and poorly crafted schemas.

\subsection{Setup}
\cite{kurt2023say, tam2024letspeakfreelystudy} have introduced a series of tasks to investigate potential quality concerns in constrained decoding, which we leverage and extend in this benchmark. Specifically, we adopt the three reasoning tasks from these studies to evaluate the impact of constrained decoding on task accuracy, as detailed in Table~\ref{tab:task_description}. The simple output structure of these tasks was designed to isolate the effects of constrained decoding on reasoning, as outlined by \cite{tam2024letspeakfreelystudy}.

For our experiments, we use the Llama-3.1-8B-Instruct model to measure task performance. We follow the original setup and prompt specifications from \cite{kurt2023say}, with full details provided in Appendix~\ref{app:task_experiment_details}.

\noindent
\renewcommand{\arraystretch}{1.2} %
\begin{longtable}{>{\raggedright\arraybackslash}p{0.20\textwidth}%
    >{\raggedright\arraybackslash}p{0.35\textwidth}%
    >{\raggedright\arraybackslash}p{0.18\textwidth}%
    >{\raggedright\arraybackslash}p{0.15\textwidth}}
\caption{Task Descriptions and Structures}
\label{tab:task_description} \\
\hline
\textbf{Task} & \textbf{Example} & \textbf{Structure} & \textbf{Metric} \\
\hline
\endfirsthead
\hline
\textbf{Task} & \textbf{Example} & \textbf{Structure} & \textbf{Metric} \\
\hline
\endhead
\hline
\endfoot
\endlastfoot

\textbf{Last Letter} & \textbf{Input:} Ian Peter Bernard Stephen \newline \textbf{Output:} nrdn & CoT reasoning + answer in $a-z$ & Case-sensitive exact match \\
\hline
\textbf{Shuffle Objects} & \textbf{Input:} Sequence of exchanges among individuals + choices\newline \textbf{Output:} A-E & CoT reasoning + answer in $A-E$ & Exact match \\
\hline
\textbf{GSM8K} & \textbf{Input:} Basic caculation problems \newline \textbf{Output:} Number, e.g. 8 & CoT reasoning + answer as integer & Number exact match \\
\hline
\end{longtable}

We implement the following constraints for the first three tasks: \textbf{(1) Last Letter} the output needs to be a concatenation of letters from a-z; \textbf{(2) Shuffle Objects} the output needs to be a single letter from A-E enclosed in parentheses; \textbf{(3) GSM8K} the output is an valid integer or float number.
The outputs for all three tasks are structured as JSON objects with two fields: \texttt{"reasoning"} and \texttt{"answer"}, formatted as \texttt{\{"reasoning": <reasoning about the answer>, "answer": <final answer>\}}.

\subsection{Results}

The results in Table~\ref{tab:quality} show that the constrained decoding, regardless of the framework, achieves higher performance than the unconstrained setting.
Among the frameworks evaluated, Guidance consistently delivers the best performance across all tasks, with approximately a 3\% improvement over the LM-only approach in every task.
We believe this may be attributed to its token-healing implementation~\citep{guidance2024tokenhealing}.

\begin{table}[H]
    \centering
    \renewcommand{\arraystretch}{0.9}
    \caption{Performance Percentages for Various Models}
    \begin{tabular}{@{}lccc@{}}
    \toprule
    & \textbf{Last Letters} & \textbf{Shuffle Objects} & \textbf{GSM8K} \\ \midrule
    \textbf{LM only}   & 50.7\%                & 52.6\%                   & 80.1\%                 \\
    \textbf{XGrammar}     & 51.2\%                & 52.7\%                   & 83.7\%                 \\
    \textbf{Llamacpp}       & 52.0\%                & 52.6\%                   & 82.4\%                 \\
    \textbf{Outlines}       & 53.3\%                & 53.0\%                   & 81.6\%                 \\
    \textbf{Guidance}       & \best{54.0\%}         & \best{55.9\%}          & \best{83.8}\%         \\ 
    \bottomrule
    \end{tabular}
    \label{tab:quality}
\end{table}

\section{Conclusion}\label{sec:conclusion}

We have proposed a comprehensive evaluation framework for constrained decoding frameworks with JSON schemas, focusing on efficiency, coverage, and output quality. We introduced JSONSchemaBench, a benchmark comprising 10K real-world JSON schemas, to enable robust assessment under realistic conditions. Our evaluation highlights both the advancements and limitations of current state-of-the-art constrained decoding frameworks. We hope that our findings and benchmark will inform future research in structured generation and provide valuable insights to help the community identify the most effective tools and to extend capabilities with constrained decoding.

\paragraph{Acknowledgements}

We would like to express our gratitude to Junda Chen (UCSD), Paul Koch (Microsoft), and Shuqi Wang (EPFL) for their valuable help and insightful discussions, Ana-Maria Indreias (EPFL) for her assistance with data visualization, and Zheng Zhou (independent researcher) for resolving GPU-related issues.

\bibliography{custom}

\begin{thebibliography}{39}
\providecommand{\natexlab}[1]{#1}
\providecommand{\url}[1]{\texttt{#1}}
\expandafter\ifx\csname urlstyle\endcsname\relax
  \providecommand{\doi}[1]{doi: #1}\else
  \providecommand{\doi}{doi: \begingroup \urlstyle{rm}\Url}\fi

\bibitem[Analytics(2022)]{snowplow2022iglu}
Snowplow Analytics.
\newblock Iglu central.
\newblock \url{https://github.com/snowplow/iglucentral}, 2022.
\newblock Commit hash 726168e. Retrieved 19 September 2022.

\bibitem[Attouche et~al.(2022)Attouche, Baazizi, Colazzo, Ghelli, Sartiani, and
  Scherzinger]{attouche_witness_2022}
Lyes Attouche, Mohamed-Amine Baazizi, Dario Colazzo, Giorgio Ghelli, Carlo
  Sartiani, and Stefanie Scherzinger.
\newblock Witness {Generation} for {JSON} {Schema}.
\newblock \emph{Proceedings of the VLDB Endowment}, 15\penalty0 (13):\penalty0
  4002--4014, September 2022.
\newblock ISSN 2150-8097.
\newblock \doi{10.14778/3565838.3565852}.
\newblock URL \url{https://dl.acm.org/doi/10.14778/3565838.3565852}.

\bibitem[Baazizi et~al.(2021)Baazizi, Colazzo, Ghelli, Sartiani, and
  Scherzinger]{schema_corpus}
Mohamed~Amine Baazizi, Dario Colazzo, Giorgio Ghelli, Carlo Sartiani, and
  Stefanie Scherzinger.
\newblock A json schema corpus, 2021.
\newblock \url{https://github.com/sdbs-uni-p/json-schema-corpus}.

\bibitem[Berman(2025)]{python-jsonschema}
Julian Berman.
\newblock python-jsonschema.
\newblock \url{https://github.com/python-jsonschema/jsonschema}, 2025.
\newblock URL \url{https://github.com/python-jsonschema/jsonschema}.
\newblock Accessed: 2025-01-05.

\bibitem[Beurer-Kellner et~al.(2023)Beurer-Kellner, Fischer, and
  Vechev]{beurer-kellner_prompting_2023}
Luca Beurer-Kellner, Marc Fischer, and Martin Vechev.
\newblock Prompting {Is} {Programming}: {A} {Query} {Language} for {Large}
  {Language} {Models}.
\newblock \emph{Proceedings of the ACM on Programming Languages}, 7\penalty0
  (PLDI):\penalty0 1946--1969, June 2023.
\newblock ISSN 2475-1421.
\newblock \doi{10.1145/3591300}.
\newblock URL \url{http://arxiv.org/abs/2212.06094}.
\newblock arXiv:2212.06094 [cs].

\bibitem[Bowtie(2025)]{bowtie}
Bowtie.
\newblock Bowtie: A meta-validator of the json schema specification, 2025.
\newblock URL \url{https://github.com/bowtie-json-schema/bowtie/}.
\newblock DOI: 10.5281/zenodo.14646449.

\bibitem[Deutsch et~al.(2019)Deutsch, Upadhyay, and
  Roth]{deutsch_general-purpose_2019}
Daniel Deutsch, Shyam Upadhyay, and Dan Roth.
\newblock A {General}-{Purpose} {Algorithm} for {Constrained} {Sequential}
  {Inference}.
\newblock In \emph{Proceedings of the 23rd {Conference} on {Computational}
  {Natural} {Language} {Learning} ({CoNLL})}, pp.\  482--492, Hong Kong, China,
  2019. Association for Computational Linguistics.
\newblock \doi{10.18653/v1/K19-1045}.
\newblock URL \url{https://www.aclweb.org/anthology/K19-1045}.

\bibitem[Dong et~al.(2024)Dong, Ruan, Cai, Lai, Xu, Zhao, and
  Chen]{dong_xgrammar_2024}
Yixin Dong, Charlie~F. Ruan, Yaxing Cai, Ruihang Lai, Ziyi Xu, Yilong Zhao, and
  Tianqi Chen.
\newblock {XGrammar}: {Flexible} and {Efficient} {Structured} {Generation}
  {Engine} for {Large} {Language} {Models}, November 2024.
\newblock URL \url{http://arxiv.org/abs/2411.15100}.
\newblock arXiv:2411.15100 [cs].

\bibitem[Geng et~al.(2023)Geng, Josifoski, Peyrard, and
  West]{geng-etal-2023-grammar}
Saibo Geng, Martin Josifoski, Maxime Peyrard, and Robert West.
\newblock Grammar-constrained decoding for structured {NLP} tasks without
  finetuning.
\newblock In Houda Bouamor, Juan Pino, and Kalika Bali (eds.),
  \emph{Proceedings of the 2023 Conference on Empirical Methods in Natural
  Language Processing}, pp.\  10932--10952, Singapore, December 2023.
  Association for Computational Linguistics.
\newblock \doi{10.18653/v1/2023.emnlp-main.674}.
\newblock URL \url{https://aclanthology.org/2023.emnlp-main.674}.

\bibitem[Geng et~al.(2024)Geng, Gambhir, Wendler, and
  West]{geng2024bytebpetokenizationinverse}
Saibo Geng, Sankalp Gambhir, Chris Wendler, and Robert West.
\newblock Byte bpe tokenization as an inverse string homomorphism, 2024.
\newblock URL \url{https://arxiv.org/abs/2412.03160}.

\bibitem[Gerganov \& al.(2023)Gerganov and al.]{llama_cpp}
Georgi Gerganov and al.
\newblock Llama.cpp: A port of facebook's llama model in c++.
\newblock \url{https://github.com/ggerganov/llama.cpp}, 2023.
\newblock Accessed: 2025-01-16.

\bibitem[{GlaiveAI}(2024)]{glaiveai2024functioncalling}
{GlaiveAI}.
\newblock Glaive function calling dataset.
\newblock
  \url{https://huggingface.co/datasets/glaiveai/glaive-function-calling}, 2024.
\newblock Accessed: 2024-12-21.

\bibitem[Grattafiori et~al.(2024)Grattafiori, Dubey, Jauhri, Pandey, Kadian,
  Al-Dahle, Letman, Mathur, Schelten, and
  et~al.]{grattafiori2024llama3herdmodels}
Aaron Grattafiori, Abhimanyu Dubey, Abhinav Jauhri, Abhinav Pandey, Abhishek
  Kadian, Ahmad Al-Dahle, Aiesha Letman, Akhil Mathur, Alan Schelten, and
  Alex~Vaughan et~al.
\newblock The llama 3 herd of models, 2024.
\newblock URL \url{https://arxiv.org/abs/2407.21783}.

\bibitem[{Guidance AI}(2023)]{guidance}
{Guidance AI}.
\newblock Guidance: A language model programming framework, 2023.
\newblock URL \url{https://github.com/guidance-ai/guidance}.
\newblock Accessed: 2024-12-18.

\bibitem[{GuidanceAI}(2024{\natexlab{a}})]{guidance2024tokenhealing}
{GuidanceAI}.
\newblock Prompt boundaries and token healing.
\newblock
  \url{https://github.com/guidance-ai/guidance/blob/main/notebooks/art_of_prompt_design/prompt_boundaries_and_token_healing.ipynb},
  2024{\natexlab{a}}.
\newblock Accessed: 2024-12-21.

\bibitem[{GuidanceAI}(2024{\natexlab{b}})]{guidance_acceleration}
{GuidanceAI}.
\newblock Guidance acceleration tutorial.
\newblock
  \url{https://guidance.readthedocs.io/en/stable/example_notebooks/tutorials/guidance_acceleration.html},
  2024{\natexlab{b}}.
\newblock Accessed: 2025-01-16.

\bibitem[{JSON Schema Org}(2024)]{jsonSchemaTestSuite}
{JSON Schema Org}.
\newblock Json schema test suite.
\newblock \url{https://github.com/json-schema-org/JSON-Schema-Test-Suite},
  2024.
\newblock URL \url{https://github.com/json-schema-org/JSON-Schema-Test-Suite}.
\newblock Accessed: 2024-12-19.

\bibitem[Kubernetes(2022)]{kubernetes2022schemas}
Kubernetes.
\newblock Kubernetes json schemas.
\newblock \url{https://github.com/instrumenta/kubernetes-json-schema}, 2022.
\newblock Commit hash 133f848.

\bibitem[Kuchnik et~al.(2023)Kuchnik, Smith, and
  Amvrosiadis]{kuchnik_validating_2023}
Michael Kuchnik, Virginia Smith, and George Amvrosiadis.
\newblock Validating {Large} {Language} {Models} with {ReLM}, May 2023.
\newblock URL \url{http://arxiv.org/abs/2211.15458}.
\newblock arXiv:2211.15458 [cs].

\bibitem[Kurt(2024{\natexlab{a}})]{coalescence2024}
Will Kurt.
\newblock Coalescence: Making llm inference 5x faster.
\newblock \url{https://blog.dottxt.co/coalescence.html}, 2024{\natexlab{a}}.
\newblock Accessed: 2024-12-21.

\bibitem[Kurt(2024{\natexlab{b}})]{kurt2023say}
Will Kurt.
\newblock Say what you mean: A response to 'let me speak freely',
  2024{\natexlab{b}}.
\newblock URL
  \url{https://.txt.co/blog/say-what-you-mean-a-response-to-let-me-speak-freely}.

\bibitem[Liu et~al.(2024)Liu, Liu, Fiannaca, Koo, Dixon, Terry, and
  Cai]{liu_we_2024}
Michael~Xieyang Liu, Frederick Liu, Alexander~J. Fiannaca, Terry Koo, Lucas
  Dixon, Michael Terry, and Carrie~J. Cai.
\newblock "{We} {Need} {Structured} {Output}": {Towards} {User}-centered
  {Constraints} on {Large} {Language} {Model} {Output}.
\newblock In \emph{Extended {Abstracts} of the {CHI} {Conference} on {Human}
  {Factors} in {Computing} {Systems}}, pp.\  1--9, May 2024.
\newblock \doi{10.1145/3613905.3650756}.
\newblock URL \url{http://arxiv.org/abs/2404.07362}.
\newblock arXiv:2404.07362 [cs].

\bibitem[Poesia et~al.(2022)Poesia, Polozov, Le, Tiwari, Soares, Meek, and
  Gulwani]{poesia_synchromesh_2022}
Gabriel Poesia, Oleksandr Polozov, Vu~Le, Ashish Tiwari, Gustavo Soares,
  Christopher Meek, and Sumit Gulwani.
\newblock Synchromesh: {Reliable} code generation from pre-trained language
  models, January 2022.
\newblock URL \url{http://arxiv.org/abs/2201.11227}.
\newblock arXiv:2201.11227 [cs].

\bibitem[Polak \& Morgan(2024)Polak and Morgan]{Polak_2024}
Maciej~P. Polak and Dane Morgan.
\newblock Extracting accurate materials data from research papers with
  conversational language models and prompt engineering.
\newblock \emph{Nature Communications}, 15\penalty0 (1), February 2024.
\newblock ISSN 2041-1723.
\newblock \doi{10.1038/s41467-024-45914-8}.
\newblock URL \url{http://dx.doi.org/10.1038/s41467-024-45914-8}.

\bibitem[Post(2022)]{washingtonpost2022schema}
The~Washington Post.
\newblock ans-schema.
\newblock \url{https://github.com/washingtonpost/ans-schema}, 2022.
\newblock Commit hash abdd6c211. Retrieved 19 September 2022.

\bibitem[Roy et~al.(2024)Roy, Thomson, Chen, Shin, Pauls, Eisner, and
  Durme]{roy_benchclamp_2024}
Subhro Roy, Sam Thomson, Tongfei Chen, Richard Shin, Adam Pauls, Jason Eisner,
  and Benjamin~Van Durme.
\newblock {BenchCLAMP}: {A} {Benchmark} for {Evaluating} {Language} {Models} on
  {Syntactic} and {Semantic} {Parsing}, January 2024.
\newblock URL \url{http://arxiv.org/abs/2206.10668}.
\newblock arXiv:2206.10668 [cs].

\bibitem[{Schema Store Org}(2014)]{jsonschemastore}
{Schema Store Org}.
\newblock The largest collection of independent json schemas in the world.
\newblock \url{https://www.schemastore.org/json/}, 2014.
\newblock A universal JSON schema store where schemas for popular JSON
  documents can be found. Contributions are welcome; see CONTRIBUTING.md for
  more information.

\bibitem[Schick et~al.(2023)Schick, Dwivedi-Yu, Dessi, Raileanu, Lomeli,
  Hambro, Zettlemoyer, Cancedda, and Scialom]{schick_toolformer_2023}
Timo Schick, Jane Dwivedi-Yu, Roberto Dessi, Roberta Raileanu, Maria Lomeli,
  Eric Hambro, Luke Zettlemoyer, Nicola Cancedda, and Thomas Scialom.
\newblock Toolformer: {Language} {Models} {Can} {Teach} {Themselves} to {Use}
  {Tools}.
\newblock In A.~Oh, T.~Naumann, A.~Globerson, K.~Saenko, M.~Hardt, and
  S.~Levine (eds.), \emph{Advances in {Neural} {Information} {Processing}
  {Systems}}, volume~36, pp.\  68539--68551. Curran Associates, Inc., 2023.
\newblock URL
  \url{https://proceedings.neurips.cc/paper_files/paper/2023/file/d842425e4bf79ba039352da0f658a906-Paper-Conference.pdf}.

\bibitem[Scholak et~al.(2021)Scholak, Schucher, and
  Bahdanau]{scholak_picard_2021}
Torsten Scholak, Nathan Schucher, and Dzmitry Bahdanau.
\newblock {PICARD}: Parsing incrementally for constrained auto-regressive
  decoding from language models.
\newblock In Marie-Francine Moens, Xuanjing Huang, Lucia Specia, and Scott
  Wen-tau Yih (eds.), \emph{Proceedings of the 2021 Conference on Empirical
  Methods in Natural Language Processing}, pp.\  9895--9901, Online and Punta
  Cana, Dominican Republic, November 2021. Association for Computational
  Linguistics.
\newblock \doi{10.18653/v1/2021.emnlp-main.779}.
\newblock URL \url{https://aclanthology.org/2021.emnlp-main.779/}.

\bibitem[Shin et~al.(2021)Shin, Lin, Thomson, Chen, Roy, Platanios, Pauls,
  Klein, Eisner, and Van~Durme]{shin_constrained_2021}
Richard Shin, Christopher Lin, Sam Thomson, Charles Chen, Subhro Roy,
  Emmanouil~Antonios Platanios, Adam Pauls, Dan Klein, Jason Eisner, and
  Benjamin Van~Durme.
\newblock Constrained {Language} {Models} {Yield} {Few}-{Shot} {Semantic}
  {Parsers}.
\newblock In \emph{Proceedings of the 2021 {Conference} on {Empirical}
  {Methods} in {Natural} {Language} {Processing}}, pp.\  7699--7715, Online and
  Punta Cana, Dominican Republic, November 2021. Association for Computational
  Linguistics.
\newblock \doi{10.18653/v1/2021.emnlp-main.608}.
\newblock URL \url{https://aclanthology.org/2021.emnlp-main.608}.

\bibitem[Tam et~al.(2024)Tam, Wu, Tsai, Lin, Lee, and
  Chen]{tam2024letspeakfreelystudy}
Zhi~Rui Tam, Cheng-Kuang Wu, Yi-Lin Tsai, Chieh-Yen Lin, Hung-yi Lee, and
  Yun-Nung Chen.
\newblock Let me speak freely? a study on the impact of format restrictions on
  large language model performance.
\newblock In Franck Dernoncourt, Daniel Preo{\c{t}}iuc-Pietro, and Anastasia
  Shimorina (eds.), \emph{Proceedings of the 2024 Conference on Empirical
  Methods in Natural Language Processing: Industry Track}, pp.\  1218--1236,
  Miami, Florida, US, November 2024. Association for Computational Linguistics.
\newblock \doi{10.18653/v1/2024.emnlp-industry.91}.
\newblock URL \url{https://aclanthology.org/2024.emnlp-industry.91/}.

\bibitem[Tang et~al.(2024)Tang, Zong, Phang, Zhao, Zhou, Cohan, and
  Gerstein]{tang_struc-bench_2024}
Xiangru Tang, Yiming Zong, Jason Phang, Yilun Zhao, Wangchunshu Zhou, Arman
  Cohan, and Mark Gerstein.
\newblock Struc-{Bench}: {Are} {Large} {Language} {Models} {Good} at
  {Generating} {Complex} {Structured} {Tabular} {Data}?
\newblock In Kevin Duh, Helena Gomez, and Steven Bethard (eds.),
  \emph{Proceedings of the 2024 {Conference} of the {North} {American}
  {Chapter} of the {Association} for {Computational} {Linguistics}: {Human}
  {Language} {Technologies} ({Volume} 2: {Short} {Papers})}, pp.\  12--34,
  Mexico City, Mexico, June 2024. Association for Computational Linguistics.
\newblock \doi{10.18653/v1/2024.naacl-short.2}.
\newblock URL \url{https://aclanthology.org/2024.naacl-short.2}.

\bibitem[Vivien(2024)]{vivien2024regexconstraints}
Vivien.
\newblock Llm decoding with regex constraints.
\newblock
  \url{https://vivien000.github.io/blog/journal/llm-decoding-with-regex-constraints.html},
  2024.
\newblock Accessed: 2024-12-21.

\bibitem[Wang et~al.()Wang, Wang, Wang, Cao, Saurous, and
  Kim]{wang_grammar_nodate}
Bailin Wang, Zi~Wang, Xuezhi Wang, Yuan Cao, Rif~A Saurous, and Yoon Kim.
\newblock Grammar {Prompting} for {Domain}-{Specific} {Language} {Generation}
  with {Large} {Language} {Models}.

\bibitem[Willard \& Louf(2023)Willard and Louf]{willard_efficient_2023}
Brandon~T. Willard and Rémi Louf.
\newblock Efficient {Guided} {Generation} for {Large} {Language} {Models},
  August 2023.
\newblock URL \url{http://arxiv.org/abs/2307.09702}.
\newblock arXiv:2307.09702 [cs].

\bibitem[Wright et~al.(2022)Wright, Andrews, Hutton, and
  Dennis]{jsonschema2020draft}
Austin Wright, Henry Andrews, Ben Hutton, and Greg Dennis.
\newblock Draft 2020-12: Json schema core specification.
\newblock \url{https://json-schema.org/draft/2020-12/json-schema-core.html},
  2022.
\newblock Published 16 June 2022. Metaschema available at
  \url{https://json-schema.org/draft/2020-12/schema}.

\bibitem[Yao et~al.(2023{\natexlab{a}})Yao, Chen, Hanjie, Yang, and
  Narasimhan]{yao_collie_2023}
Shunyu Yao, Howard Chen, Austin~W. Hanjie, Runzhe Yang, and Karthik Narasimhan.
\newblock {COLLIE}: {Systematic} {Construction} of {Constrained} {Text}
  {Generation} {Tasks}, July 2023{\natexlab{a}}.
\newblock URL \url{http://arxiv.org/abs/2307.08689}.
\newblock arXiv:2307.08689 [cs].

\bibitem[Yao et~al.(2023{\natexlab{b}})Yao, Zhao, Yu, Du, Shafran, Narasimhan,
  and Cao]{yao2023reactsynergizingreasoningacting}
Shunyu Yao, Jeffrey Zhao, Dian Yu, Nan Du, Izhak Shafran, Karthik Narasimhan,
  and Yuan Cao.
\newblock {ReAct}: Synergizing reasoning and acting in language models.
\newblock In \emph{International Conference on Learning Representations
  (ICLR)}, 2023{\natexlab{b}}.

\bibitem[Zheng et~al.(2024)Zheng, Yin, Xie, Sun, Huang, Yu, Cao, Kozyrakis,
  Stoica, Gonzalez, Barrett, and Sheng]{zheng_sglang_2024}
Lianmin Zheng, Liangsheng Yin, Zhiqiang Xie, Chuyue Sun, Jeff Huang, Cody~Hao
  Yu, Shiyi Cao, Christos Kozyrakis, Ion Stoica, Joseph~E. Gonzalez, Clark
  Barrett, and Ying Sheng.
\newblock {SGLang}: {Efficient} {Execution} of {Structured} {Language} {Model}
  {Programs}, June 2024.
\newblock URL \url{http://arxiv.org/abs/2312.07104}.
\newblock arXiv:2312.07104 [cs].

\end{thebibliography}
\bibliographystyle{iclr2025_conference}

\appendix

\section{JSON Schema Collections Details}\label{app:schema_collection_details}

JSONSchemaBench includes a diverse collection of schemas curated from multiple real-world applications\citet{attouche_witness_2022}, designed to represent a wide range of use cases:

\paragraph{Sources:}
\begin{itemize}
    \item \textbf{GitHub~\citep{schema_corpus}}: Extracted from open-source repositories containing schema definitions, representing practical, widely-used applications. Schemas from GitHub are of various complexities, totaling 6,000 schemas. We split the collection into trivial (fewer than 10 fields), easy (10–30 fields), medium (30–100 fields), hard (100–500 fields), and ultra (more than 500 fields), based on the total number of fields in each JSON schema to reflect increasing complexity and scale.
    \item \textbf{Snowplow~\citep{snowplow2022iglu}}: Sourced from event-based analytics frameworks, showcasing schemas tailored for event-driven data structures.
    \item \textbf{Kubernetes~\citep{kubernetes2022schemas}}: Schemas defining configurations for container orchestration systems, highlighting schemas with intricate hierarchical structures.
    \item \textbf{WashingtonPost~\citep{washingtonpost2022schema}}: Schemas for The Washington Post's ANS specification.
    \item \textbf{GlaiveAI2K~\cite{glaiveai2024functioncalling}}: 2,000 schemas extracted from a function-calling dataset. Each schema represents a function signature.
    \item \textbf{JSON Schema Store~\citep{jsonschemastore}}: The largest collection of independent JSON schemas in the world.
\end{itemize}

\begin{table}[ht]
    \centering
    \caption{Baisc statistics of the datasets used in the experiments.}
    \label{tab:dataset-characteristics}
    \setlength{\tabcolsep}{3pt} %
    {\footnotesize %
    \begin{tabular}{@{}lllllll@{}}
        \toprule
        \textbf{Dataset}  & \textbf{Count} & \textbf{Size (KB)} & \textbf{Field Count} & \textbf{Max Fan-Out} & \textbf{Schema Depth}  \\ 
        & &  \textbf{Med / Max} & \textbf{Med / Max} & \textbf{Med / Max} & \textbf{Med / Max}  \\ 
        \midrule
        GlaiveAI-2K  & 1707 & 0.5 / 1.2 & 21 / 44 & 4 / 7 & 5 / 8  \\
        Github-Trivial  & 444 & 0.2 / 10.8 & 6 / 9 & 4 / 9 & 2 / 6  \\
        Github-Easy  & 1943 & 0.5 / 20.3 & 18 / 29 & 5 / 19 & 4 / 10  \\
        Snowplow  & 403 & 0.9 / 15.6 & 37 / 450 & 7 / 131 & 3 / 13  \\
        Github-Medium  & 1976 & 1.5 / 58.3 & 51 / 99 & 8 / 42 & 6 / 15  \\
        Kubernetes  & 1064 & 2.7 / 818.6 & 41 / 11720 & 5 / 600 & 5 / 7  \\
        Washington Post  & 125 & 1.7 / 81.1 & 44 / 2093 & 7 / 84 & 4 / 10  \\
        Github-Hard  & 1240 & 5.1 / 136.1 & 175 / 498 & 18 / 133 & 8 / 25  \\
        JSONSchemaStore  & 492 & 5.9 / 2934.8 & 155 / 108292 & 14 / 6543 & 6 / 22  \\
        Github-Ultra  & 164 & 25.8 / 359.6 & 694 / 6919 & 37 / 412 & 8 / 23  \\
        \bottomrule
    \end{tabular}
    }
\end{table}

\subsection{Data Processing}

To ensure the quality and reliability of JSONSchemaBench, we applied the following preprocessing steps:

\paragraph{1. Validation}
\begin{itemize}
    \item Verified schemas conform to the JSON Schema specification using the \texttt{jsonschema} library in Python, specifically targeting the \texttt{Draft2020-12} version. Drop invalid schemas.
    \item Identified additional invalid schemas using validators from Rust and JavaScript libraries.
\end{itemize}

\paragraph{2. Cleaning}
\begin{itemize}
    \item \textbf{Deduplicate:} Removed duplicate schemas to eliminate redundancy and maintain a diverse dataset. Key ordering within schemas was ignored when determining duplicates.
    \item \textbf{Empty Schema:} Excluded schemas that were lacking meaningful constraints, effectively ``empty.''
    \item \textbf{Unresolved References:} Removed schemas containing unresolved \texttt{\$ref} references to external URLs.
    \item \textbf{Schema Version Fixes:} Corrected mismatched or missing draft versions.
    \item \textbf{Extraneous Field Removal:} Eliminated unrelated fields such as \texttt{command}, \texttt{config}, \texttt{path}, and \texttt{controls}.
    \item \textbf{Regex Escaping:} Fixed escaping issues in regular expressions to ensure validity.
    \item \textbf{Schema Extraction:} Extracted schemas embedded within non-root levels of JSON files.
\end{itemize}

\subsection{Draft versions}

\begin{table}[h]
    \centering
    \caption{JSON Schema Draft Version Counts}
    \label{tab:JSONSchema_draft_count}
    \begin{tabular}{lrrrrrr}
        \toprule
         & draft-04 & draft-06 & draft-07 & 2019-09 & 2020-12 & unknown \\
        \midrule
        Github-easy & 1310 & 54 & 136 & 0 & 5 & 438 \\
        Github-hard & 841 & 30 & 87 & 0 & 23 & 259 \\
        Github-medium & 1221 & 80 & 140 & 0 & 7 & 528 \\
        JsonSchemaStore & 199 & 5 & 268 & 5 & 11 & 4 \\
        Kubernetes & 0 & 0 & 0 & 0 & 0 & 1087 \\
        Snowplow & 0 & 0 & 0 & 0 & 0 & 408 \\
        WashingtonPost & 125 & 0 & 0 & 0 & 0 & 0 \\
        Glaiveai2K & 0 & 0 & 0 & 0 & 0 & 1707 \\
        total & 4097 & 193 & 706 & 5 & 50 & 5155 \\
        \bottomrule
    \end{tabular}

\end{table}

\subsection{Feature Distribution}

\begin{figure}[H]
    \centering
    \begin{subfigure}[b]{0.65\textwidth}
        \centering
        \includegraphics[width=\textwidth]{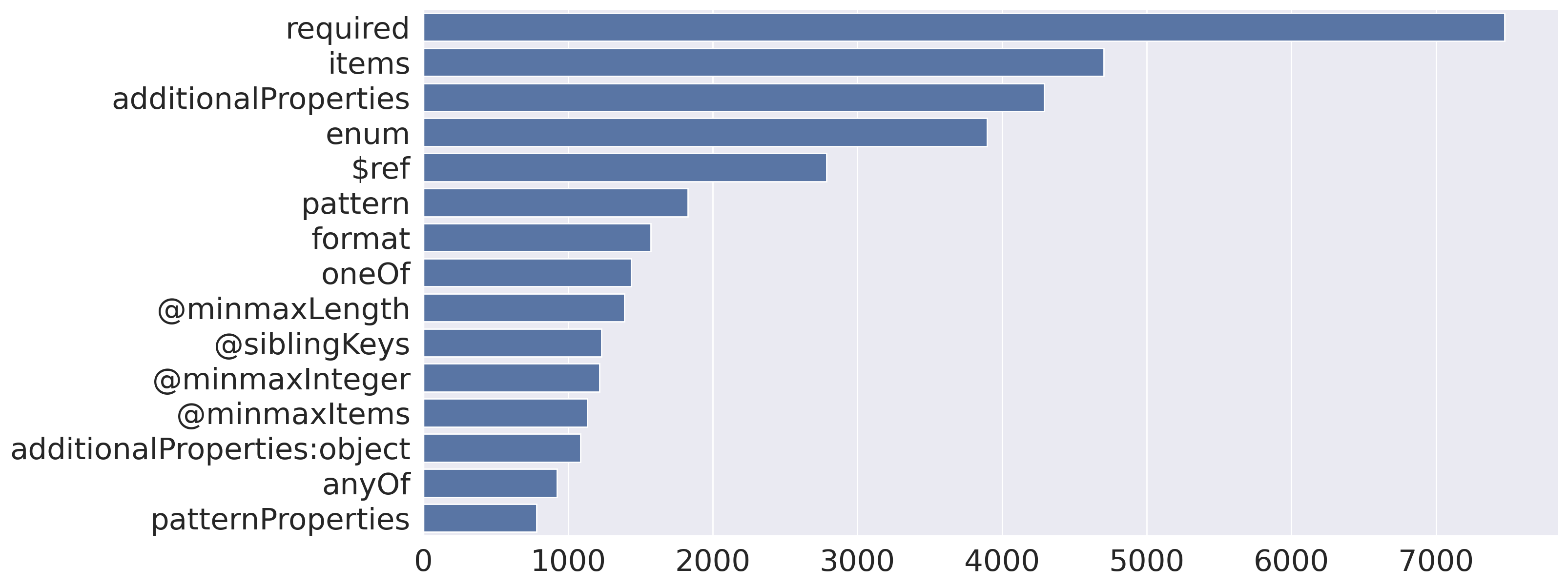}
        \caption{Feature Count in the 10K Schemas}
        \label{fig:feature_visualization}
    \end{subfigure}
    \hfill
    \begin{subfigure}[b]{0.33\textwidth}
        \centering
        \includegraphics[width=\textwidth]{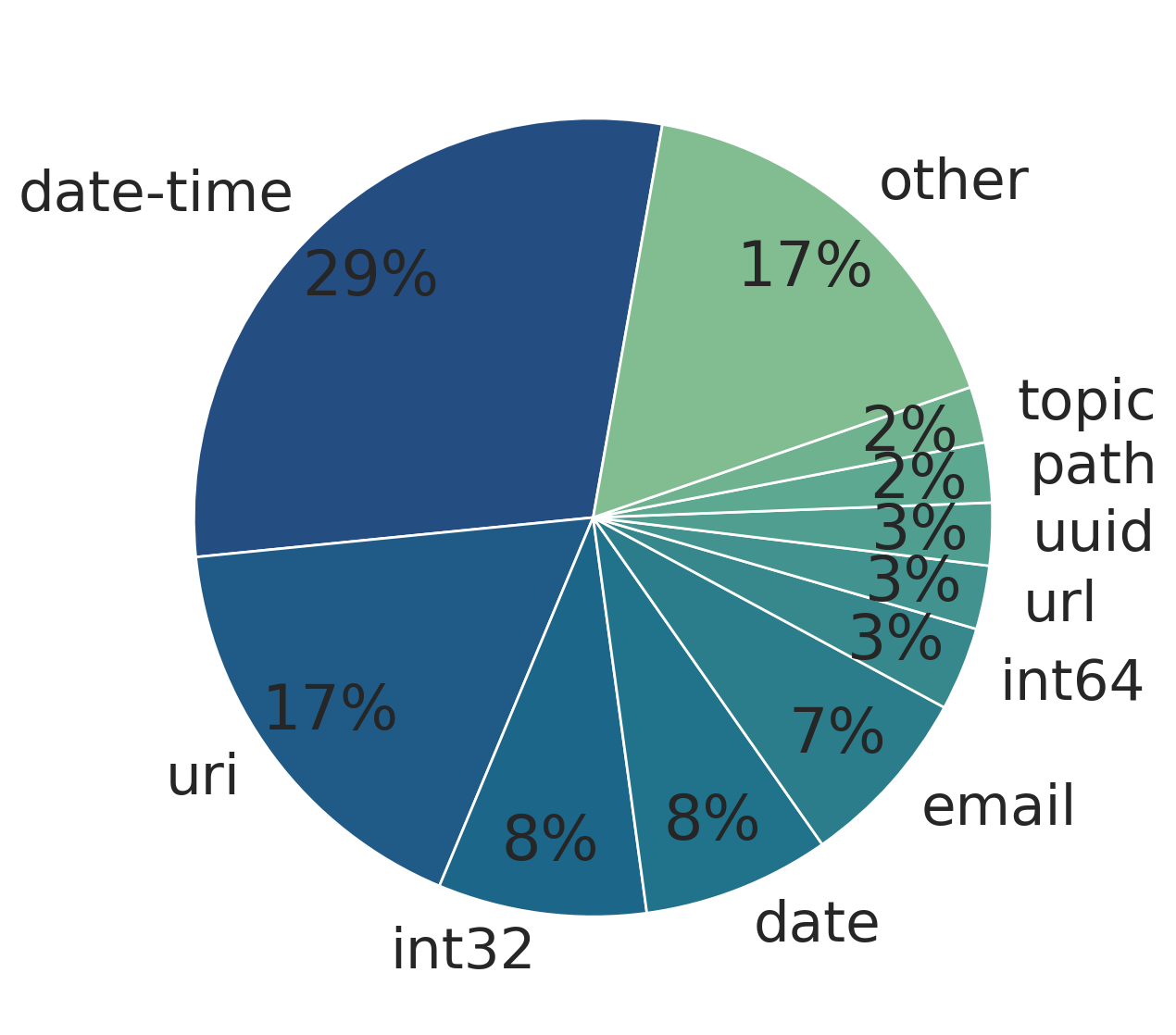}
        \caption{Format keyword distribution}
        \label{fig:format_visualization}
    \end{subfigure}
    \caption{Feature and Format constraint distribution. }
    \label{fig:feature_and_format}
\end{figure}

We count the appearance of each feature (keyword) in the 10K schemas and show the most frequent features in Figure~\ref{fig:feature_visualization}.
We separately plot usage of the \texttt{format} keyword, which is used to specify format of string such as \texttt{date-time}, \texttt{email}, \texttt{uri}. This is worth highlighted because each of these formats can be quite complex to implement on its own
The distribution of formats used is shown in Figure~\ref{fig:format_visualization}.

\section{Coverage Experiment Details}\label{app:coverage_experiment_details}

The prompting template used for the coverage experiment is shown in Figure~\ref{fig:json-prompt-template}.

\begin{figure}[H]
    \begin{tcolorbox}[
        colback=gray!5!white,
        colframe=gray!75!black,
        title=Prompt Template for JSON Generation,
        width=\textwidth,
        boxrule=0.75pt,
        arc=2mm,
        fonttitle=\bfseries
    ]
    \textbf{System Message:} \\
    You need to generate a JSON object that matches the schema below. \\

    \textbf{Demo Examples:} \\
    \verb|## Input Schema:| \texttt{\textcolor{gray}{[JSON schema]}} \\
    \verb|## Expected Output:| \texttt{\textcolor{gray}{[JSON object matching the schema]}}\\
    ...
    \end{tcolorbox}
    \caption{Prompt template used to generate JSON objects in the coverage experiment.}
    \label{fig:json-prompt-template}
\end{figure}

\paragraph{Decoding Method}

We use greedy decoding with no top P or top K sampling for all the experiments.
We only get one output from the model, which we will use to validate the schema compliance.
It's totally plausible to sample more outputs and validate them all, and it might detect more schema violations. 
The fact that we only sample the top 1 output may quantify our \textit{empirical coverage} as \textit{Top 1 Empirical Coverage}.

\paragraph{Validation}

We use the \texttt{jsonschema} library with the Draft-2020-12 version of the JSON Schema standard to validate the generated JSON object.
We turn on the `format' checks, which are not enabled by default in Python.
Strictly speaking, the \texttt{jsonschema} library doesn't guarantee the validation of all the schema constraints, even with the `format' checks enabled.
It is possible, though very rare, for a schema-noncompliant output to be validated as compliant by the \texttt{jsonschema} library, leading to a slight overestimation of empirical coverage. However, such occurrences are corner cases and happen infrequently.

\section{Theoretical Coverage Details}\label{app:theoretical_coverage_details}

\begin{definition}[Theoretical Coverage]
    A schema is considered \textit{theoretically covered} if all of its features are supported by the grammar engine. 
\end{definition}

The \textit{theoretical coverage}, noted as $\cal{C}_{\text{Theoretical}}$, measures the proportion of JSON schemas that a grammar engine supports based on its implementation. 
It doesn't involve any model inference or experiments and is solely based on the grammar engine's implementation.
$\cal{C}_{\text{Theoretical}}$ is an \textit{upper bound} of the \textit{true coverage}, which cannot be empirically measured due to the infinite number of possible generations under the schema constraints.

Overall, the theoretical coverage provides a good indication of the grammar engine's capability to support a wide range of schema constraints.

In our experiment, the theoretical coverage for each framework was determined based on the documentation and resources listed in Table~\ref{tab:grammar-engine-version}. 

\begin{table}[ht]
    \centering
    \caption{Grammar Engine Documentation and Resources}
    \small
    \begin{tabular}{@{}lllll@{}}
    \toprule
    \textbf{Frameworks} & \textbf{Lib Version} & \textbf{Release Date} & \textbf{JSON Schema Support Documentation} \\
    \midrule
    Guidance                & 0.2.0rc                 & 2024.11.26               & \href{https://github.com/guidance-ai/llguidance/blob/a7b69d6c57ac514b32afbed52a9c292cb4b4b3bd/parser/src/json/README.md}{LLGuidance Documentation} \\ 
    Llamacpp        & 0.3.2               & 2024.11.16               & \href{https://github.com/ggerganov/llama.cpp/blob/66c2c93082289325199ae1f773f3b0ab2e399a47/common/json-schema-to-grammar.cpp}{llama.cpp JSON Schema to gbnf Conversion} \\ 
    XGrammar                & 0.1.6                  & 2024.12.07                     & \href{https://github.com/mlc-ai/xgrammar/blob/5e141f6ff1ca02bc31f9e512e68b61f2a8ae88e5/cpp/json_schema_converter.cc}{XGrammar JSON Schema to gbnf Conversion} \\ 
    Outlines                & 0.1.8                & 2024.12.06               & \href{https://github.com/dottxt-ai/outlines/blob/e4f96fbce45593222e40805ca614ace251728ef2/outlines/fsm/json_schema.py}{Outlines JSON Schema to Regex Conversion} \\ 
    OpenAI                  & UNK                & UNK               & \href{https://platform.openai.com/docs/guides/structured-outputs#supported-schemas}{OpenAI Structured Output API} \\ 
    Gemini           & 0.8.3                & 2024.10.31               & \href{https://github.com/google-gemini/generative-ai-python/blob/bcb7cf968ed3b3b858e66ac85b08ba9925ba8e97/google/generativeai/types/content_types.py}{Gemini Structured Output Content Types} \\ 
    \bottomrule
    \end{tabular}
    \label{tab:grammar-engine-version}
\end{table}

\begin{figure}[ht]
    \centering
    \includegraphics[width=0.99\textwidth]{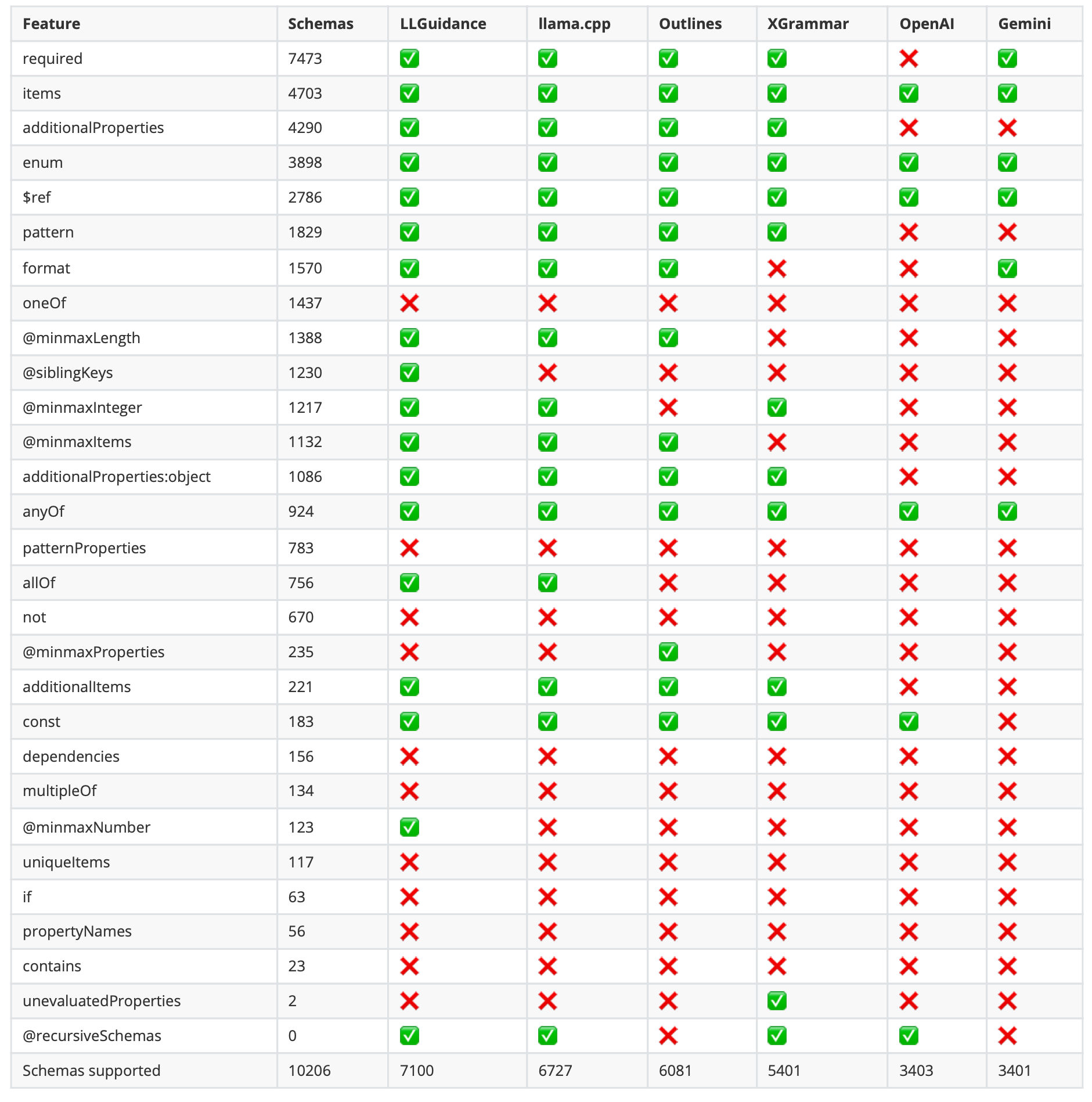}
    \caption{Feature checklist for different structured output engines}
    \label{fig:keywords_checklist}
\end{figure}

The theoretical support for each feature in JSON Schema is summarized in Figure~\ref{fig:keywords_checklist}

\begin{table}[ht]
\centering
\scriptsize
\caption{Theoretical coverage across datasets.}
\label{tab:theoretical_coverage}
\begin{tabular}{lccccccc}
\toprule
\textbf{Dataset} & \textbf{LM only} & \textbf{Guidance} & \textbf{Llamacpp} & \textbf{Outlines} & \textbf{XGrammar} & \textbf{OpenAI} & \textbf{Gemini} \\ 
\midrule
GlaiveAI        & 0.00 & \textbf{0.96} & 0.95 & 0.95 & 0.87 & 0.87 & 0.87 \\
GitHub Easy     & 0.00 & \textbf{0.87} & 0.83 & 0.75 & 0.65 & 0.31 & 0.31 \\
Snowplow        & 0.00 & \textbf{0.80} & 0.74 & 0.58 & NA   & 0.29 & NA   \\
GitHub Medium   & 0.00 & \textbf{0.73} & 0.69 & 0.57 & 0.49 & 0.22 & NA   \\
Kubernetes      & 0.00 & \textbf{0.58} & 0.58 & 0.58 & 0.58 & 0.40 & NA   \\
Washington Post & 0.00 & \textbf{0.70} & 0.64 & 0.63 & 0.62 & 0.29 & NA   \\
GitHub Hard     & 0.00 & \textbf{0.54} & 0.49 & 0.38 & 0.33 & 0.00 & NA   \\
JsonSchemaStore & 0.00 & \textbf{0.31} & 0.24 & 0.20 & 0.13 & 0.00 & NA   \\
\bottomrule
\end{tabular}

\end{table}

The theoretical coverage of each grammar engine is summarized in Table~\ref{tab:theoretical_coverage}.

\section{JSON Schema Test Suite Experiment Details}\label{app:test_suite_experiment_details}

\begin{figure}[H]
    \centering
    \includegraphics[width=0.6\textwidth]{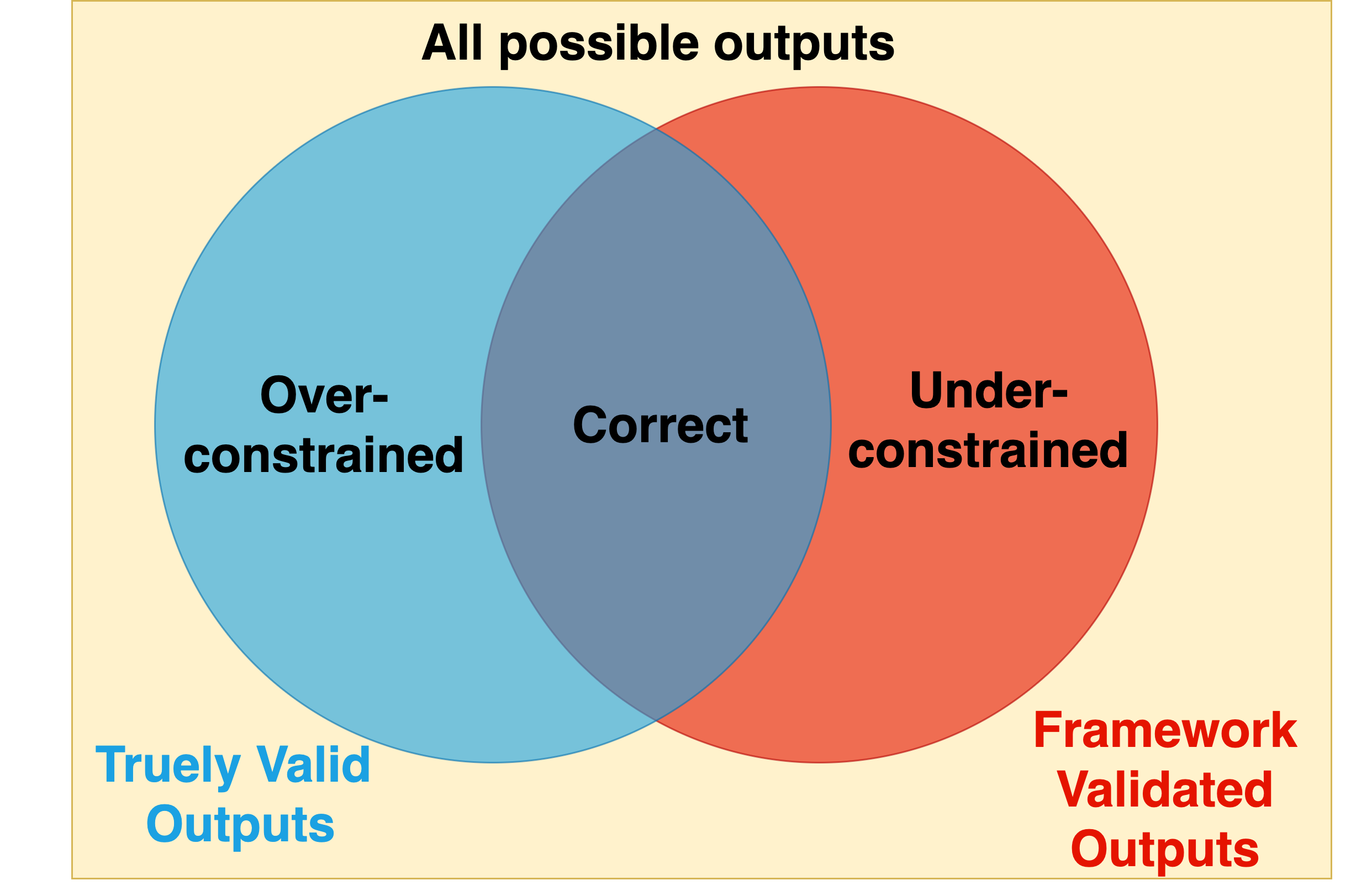} %
    \caption{Illustration of over-constrained and under-constrained.}
    \label{fig:venn_diagram_over_under}
\end{figure}

\begin{figure}[htbp]
    \centering
    \includegraphics[width=1.1\linewidth]{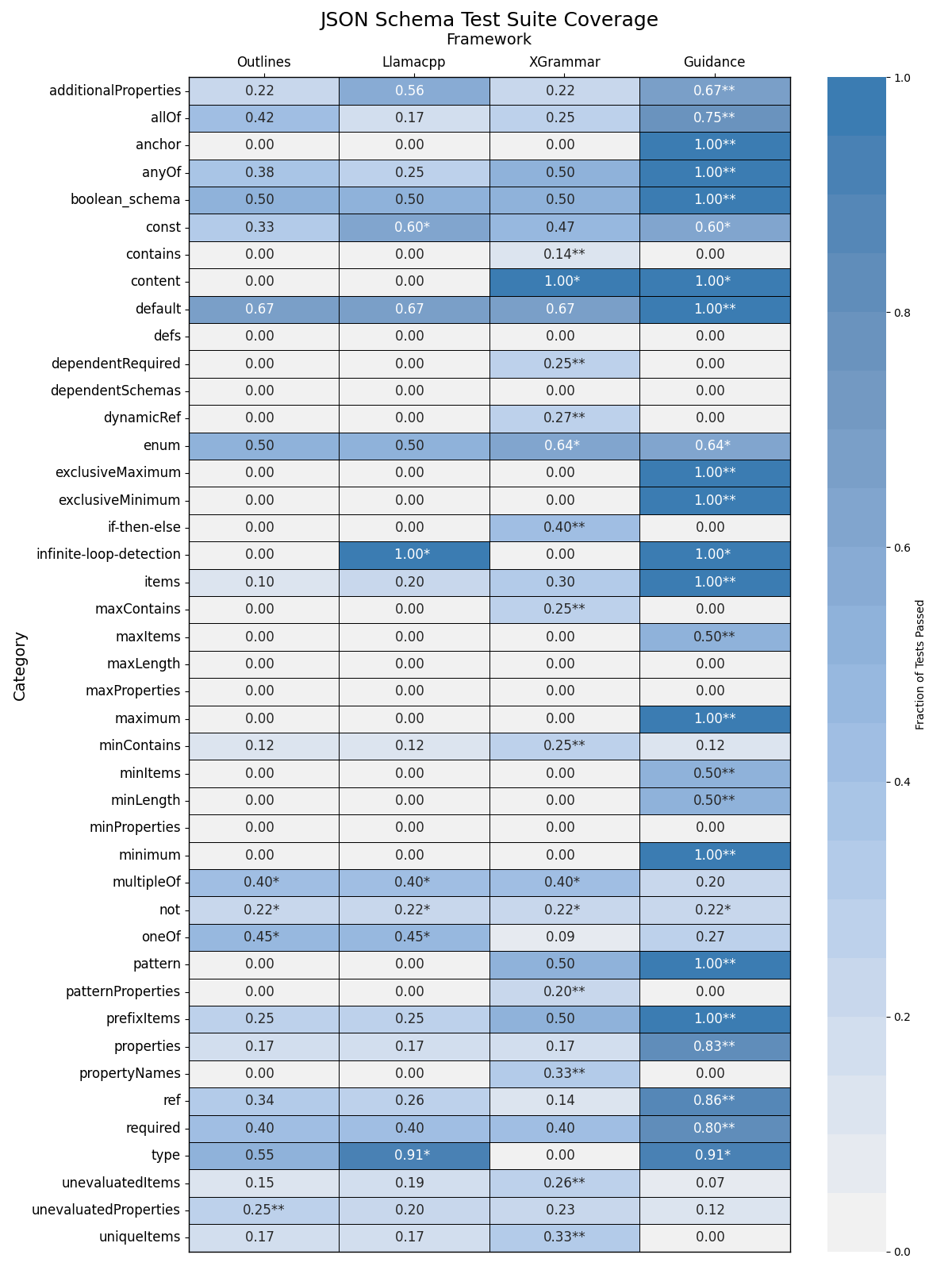}
    \caption{JSON Schema test suite coverage by category. Each cell represents the proportion of passing tests for a given category-framework pair, with darker shades indicating higher coverage. A single asterisk (*) marks frameworks tied for the highest (non-zero) coverage, while a double asterisk (**) marks the framework with the single highest coverage in the category.}
    \label{fig:test_coverage}
\end{figure}

We evaluated each constrained decoding framework's performance on the JSON Schema Test Suite using the following criteria: a framework is considered to pass a test case if it permits generating every valid instance in the test case while preventing the generation of every invalid instance. Some test cases consist exclusively of invalid instances, such as those involving \textit{unsatisfiable} schemas, i.e., schemas for which no valid instances exist. In these cases, engines raising compile-time errors were allowed to pass.

\paragraph{Cleaning}

We removed the 'format' category of tests, as the current JSON Schema standard mandates that this keyword be ignored entirely by default. The test suite comes bundled with an 'optional' set of tests, including tests for each officially recognized value of the 'format' keyword. We hope to extend this work to include these optional tests in a follow-up.

Furthermore, some tests require external resources in the form of JSON schemas available at a remote URL. We dropped these tests from the analysis, as the constrained decoding libraries discussed in the current work do not fetch these resources by default. After filtering out these tests, we are left with 43 of the original 45 test categories.

\paragraph{Implementation}

To check whether a given framework accepts or blocks the generation of a particular JSON instance, we tokenize\footnote{The particular choice of tokenizer is not particularly important, but we use the Llama 3.1 tokenizer for consistency with our other experiments.} JSON-serialized form of the instance and walk the framework's constraints forward one token at a time, essentially simulating the generation process of an LLM attempting to produce the given token sequence:

\begin{itemize}
    \item XGrammar directly expose an interface for updating the token mask after inserting a token and checking validity.
    \item Outlines does not expose a public interface for interacting with the token mask, but \texttt{outlines-core}, which \texttt{outlines} is built on top of, is easily adapted for this purpose.
    \item Similarly, Guidance does not expose a public interface for interacting with the token mask, but \texttt{llguidance}, which \texttt{guidance} is built on top of, is easily adapted for this purpose.
    \item Llamacpp does not expose this interface, but it shares a common grammar-specification language with XGrammar. We use \texttt{llamacpp} to generate GGML BNF and check token-sequence validity using \texttt{xgrammar}'s interface.
\end{itemize}

We provide code snippets that show the use of the JSON Schema Test Suite to assess the test coverage of each constrained decoding framework. For each framework, we implemented a `test harness' according to the base classes showed in listing \ref{listing:harness_base}.

Listing \ref{listing:test_case} shows the criteria for a test case to pass, which depends on all tests in the case to pass (listing \ref{listing:test}). We show the definition of \texttt{TestCase} and \texttt{Test} in listing \ref{listing:test_case_schema}.

Concrete implementations of the test harness for each framework are reported in listings \ref{listing:harness_outlines}, \ref{listing:harness_guidance}, \ref{listing:harness_xgrammar},
and \ref{listing:harness_llamacpp}.

\begin{listing}[H]
\begin{minted}{python}
class Compiler:
    def __init__(self, model_id: str):
        """
        Builds a Compiler, taking a huggingface model_id to provide
        configuration information about the model and/or tokenizer.
        """

    def compile(self, schema: str) -> Masker:
        """
        Compiles a schema into a masker used to validate a stream of
        tokens according to the schema.

        Raises an exception if the framework cannot compile the schema.
        """

class Masker:
    def advance(self, token: int):
        """
        Advances the masker by one token.

        Raises an exception if the token is not allowed by the mask.
        """

    def assert_done(self):
        """
        Asserts that the masker is either in a terminal state or will
        accept an EOS token, after which it will be in a terminal state.
    
        Raises an exception if otherwise.
        """
\end{minted}
    \caption{Abstract test harness}
    \label{listing:harness_base}
\end{listing}

\begin{listing}[H]
\begin{minted}{python}
def do_test_case(test_case: TestCase, compiler: Compiler, tokenizer: Tokenizer) -> bool:
    try:
        masker = compiler.compile(json.dumps(test_case.schema))
    except:
        if all(not test.valid for test in test_case.tests):
            # Pass: compile error on a case with only invalid test data
            return True
        else:
            # Fail: compile error but schema has at least one valid test datum
            return False

    for test in test_case.tests:
        passed = do_test(test, tokenizer, masker.copy())
        if not passed:
            # Fail: a test failed
            return False
    # Pass: all tests passed
    return True
\end{minted}
    \caption{Running a test case}
    \label{listing:test_case}
\end{listing}

\begin{listing}[H]
\begin{minted}{python}
def do_test(test: Test, tokenizer: Tokenizer, masker: Masker) -> bool: 
    tokens = tokenizer(json.dumps(test.data), add_special_tokens=False)["input_ids"]
    try:
        for token in tokens:
            masker.advance(token)
        masker.assert_done()
    except:
        if test.valid:
            # Fail: valid data was rejected
            return False
        else:
            # Pass: invalid data was rejected
            return True
    else:
        if test.valid:
            # Pass: valid data was accepted
            return True
        else:
            # Fail: invalid data was accepted
            return False
\end{minted}
    \caption{Running a test}
    \label{listing:test}
\end{listing}

\begin{listing}[H]
\begin{minted}{python}
from pydantic import BaseModel
from typing import Any, Union

class TestCase(BaseModel):
    schema: Union[bool, dict]
    tests: list[Test]

class Test(BaseModel):
    data: Any
    valid: bool
\end{minted}
    \caption{TestCase specification}
    \label{listing:test_case_schema}
\end{listing}

\begin{listing}[H]
\begin{minted}{python}
import outlines
import outlines_core


class OutlinesCompiler(Compiler):
    def __init__(self, model_id: str):
        self.tokenizer = outlines.models.transformers(model_id).tokenizer

    def compile(self, schema: str) -> "OutlinesMasker":
        regex = build_regex_from_schema(schema)
        guide = outlines.fsm.guide.RegexGuide.from_regex(
            regex, self.tokenizer
        )
        return OutlinesMasker(guide, eos_token_id=self.tokenizer.eos_token_id)


class OutlinesMasker(Masker):
    def __init__(self, guide, eos_token_id=None):
        self.guide = guide
        self.state = self.guide.initial_state
        self.eos_token_id = eos_token_id

    def advance(self, token: int):
        assert token in self.guide.get_next_instruction(self.state).tokens
        self.state = self.guide.get_next_state(self.state, token)

    def assert_done(self):
        if not self.guide.is_final_state(self.state):
            assert self.eos_token_id in self.guide.get_next_instruction(self.state).tokens
            self.advance(self.eos_token_id)
        assert self.guide.is_final_state(self.state)
\end{minted}
    \caption{Concrete test harness for Outlines}
    \label{listing:harness_outlines}
\end{listing}

\begin{listing}[H]
\begin{minted}{python}
import guidance
import llguidance

class GuidanceCompiler(Compiler):
    def __init__(self, model_id: str):
        self.gtokenizer = guidance.models.transformers.TransformersTokenizer(model_id, None)
        self.lltokenizer = llguidance.LLTokenizer(llguidance.TokenizerWrapper(self.gtokenizer))

    def compile(self, schema: str) -> GuidanceMasker:
        grammar = guidance.json(schema=schema)
        llinterpreter = llguidance.LLInterpreter(
            tokenizer=self.lltokenizer,
            llguidance_json=json.dumps(grammar.ll_serialize()),
            enable_backtrack=False,
            enable_ff_tokens=False,
        )
        return GuidanceMasker(llinterpreter, self.gtokenizer.eos_token_id)

class GuidanceMasker(Masker):
    def __init__(self, llinterpreter, eos_token_id):
        self.llinterpreter = llinterpreter
        self.eos_token_id = eos_token_id

    def advance(self, token: int):
        bytemask, _ = self.llinterpreter.compute_mask()
        assert bytemask[token] > 0
        self.llinterpreter.commit_token(token)

    def assert_done(self):
        if self.llinterpreter.stop_reason() == "NotStopped":
            bytemask, _ = self.llinterpreter.compute_mask()
            if bytemask is not None:
                assert bytemask[self.eos_token_id] > 0
                self.llinterpreter.commit_token(self.eos_token_id)
                bytemask, _ = self.llinterpreter.compute_mask()
                assert bytemask is None
        assert self.llinterpreter.stop_reason() in {"NoExtension", "EndOfSentence"}
\end{minted}
    \caption{Concrete test harness for Guidance}
    \label{listing:harness_guidance}
\end{listing}

\begin{listing}[H]
\begin{minted}{python}
import xgrammar as xgr
from transformers import AutoConfig, AutoTokenizer

class XGrammarCompiler(Compiler):
    def __init__(self, model_id: str):
        tokenizer = AutoTokenizer.from_pretrained(model_id)
        config = AutoConfig.from_pretrained(model_id)
        self.eos_token_id = tokenizer.eos_token_id
        self.tokenizer_info = xgr.TokenizerInfo.from_huggingface(
            tokenizer, vocab_size=config.vocab_size
        )
        self.compiler = xgr.GrammarCompiler(
            tokenizer_info=self.tokenizer_info,
        )

    def compile(self, schema: str) -> "XGrammarMasker":
        compiled_grammar = self.compiler.compile_json_schema(schema, strict_mode=False)
        xgr_matcher = xgr.GrammarMatcher(compiled_grammar)
        return XGrammarMasker(xgr_matcher, self.eos_token_id)


class XGrammarMasker(Masker):
    def __init__(self, xgr_matcher, eos_token_id):
        self.matcher = xgr_matcher
        self.eos_token_id = eos_token_id

    def advance(self, token: int):
        assert self.matcher.accept_token(token)

    def assert_done(self):
        if not self.matcher.is_terminated():
            self.advance(self.eos_token_id)
        assert self.matcher.is_terminated()
\end{minted}
    \caption{Concrete test harness for xGrammar}
    \label{listing:harness_xgrammar}
\end{listing}

\begin{listing}[H]
\begin{minted}{python}
from llama_cpp import LlamaGrammar
import xgrammar as xgr

class LlamacppCompiler(XGrammarCompiler):

    def compile(self, schema) -> XGrammarMasker:
        grammar_bnf = LlamaGrammar.from_json_schema(schema)._grammar
        compiled_grammar = self.compiler.compile_grammar(grammar_bnf)
        xgr_matcher = xgr.GrammarMatcher(compiled_grammar)
        return XGrammarMasker(xgr_matcher, self.eos_token_id)
\end{minted}
    \caption{Concrete test harness for Llamacpp, inheriting from the XGrammar harness for all functionality after using \texttt{llamacpp} to convert the schema to GGML BNF.}
    \label{listing:harness_llamacpp}
\end{listing}

\section{Efficiency Experiment Details}\label{app:efficiency_experiment_details}

For efficiency experiments, the results depend on both the size of the model and the tokenizer's vocabulary size. We used \textbf{Llama-3.1-8B-Instruct} (quantized to Q8bit) with a 128K token vocabulary to achieve a balance between computational efficiency and model capability.

Below, we outline specific considerations related to grammar and prefix caching:
\begin{itemize}
    \item \textbf{Grammar Cache (Compilation):} Since each schema in the dataset is unique, caching grammar compilations does not offer any benefits.
    \item \textbf{Prefix Cache (LLM Inference):} We implement prefix caching during LLM inference for all cases to enhance efficiency by reusing computed results where applicable.
\end{itemize}

\begin{table}[h!]
    \centering
    \caption{\textbf{Efficiency metrics} for different engines with \textbf{LlamaCpp} as the inference engine. 
    \textbf{GCT}: Grammar Compilation Time, \textbf{TTFT}: Time to First Token, \textbf{TPOT}: Time Per Output Token, \textbf{TGT}: Total Generation Time, \textbf{FF}: Fast-Forwarded output tokens.
    Bold values indicate the smallest in each column for GCT, TTFT, TPOT, and TGT. All values are \textbf{median} of the samples.}
    \label{tab:efficiency_llamacpp_additional}
    \renewcommand{\arraystretch}{0.9}
    \begin{adjustbox}{max width=\textwidth}
    \begin{tabular}{llrrrrr}
    \toprule
    \textbf{Dataset} & \textbf{Framework} & \textbf{GCT (s)} & \textbf{TTFT (s)} & \textbf{TPOT (ms)} & \textbf{TGT (s)} & \textbf{Output Tokens (FF)} \\
    \midrule
    \textbf{GlaiveAI}        & LLM only            &    NA    &       \best{0.10}  &       15.40  &      1.08 &           64.94 (00.00)    \\
                             & Guidance        & \best{0.00} & 0.24 & \best{6.37} & \best{0.50} & 41.56 (15.70) \\
                             & Llamacpp      & 0.05 & 0.20 & 29.98 & 1.47 & 43.18 (00.00)  \\
                             & Outlines        & 3.48 & 3.65 & 30.33 & 4.84 & 40.39 (00.00)  \\
    \cmidrule(lr){1-7}
    \textbf{GitHub Easy}     & LLM only            & NA    &       \best{0.10}  &       15.83 &      0.95 &           53.91 (00.00)    \\
                             & Guidance        & \best{0.00} & 0.34 & \best{7.44} & \best{0.60} & 34.92 (10.02) \\
                             & Llamacpp      & 0.05 & 0.18 & 27.22 & 1.10 & 33.93 (00.00)  \\
                             & Outlines        & 3.71 & 3.97 & 39.78 & 5.29 & 34.19 (00.00)  \\
    \cmidrule(lr){1-7}
    \textbf{Snowplow}        & LLM only            &    NA    &       \best{0.11} &       16.23 &      1.01 &           55.31 (00.00)    \\
                             & Guidance        & \best{0.00} & 0.28 & \best{6.55} & \best{0.51} & 36.77 (14.50) \\
                             & Llamacpp      & 0.05 & 0.20 & 28.90 & 1.24 & 37.21 (00.00)  \\
                             & Outlines        & 3.91 & 4.14 & 42.66 & 5.65 & 35.65 (00.00)  \\
    \cmidrule(lr){1-7}
    \textbf{GitHub Medium}   & LLM only            &    NA    &       \best{0.20}  &       16.68 &      2.56 &          142.10 (00.00)    \\
                             & Guidance        & \best{0.01} & 0.54 & \best{7.57} & \best{1.29} & 99.66 (31.42) \\
                             & Llamacpp      & 0.06 & 0.30 & 29.08 & 2.85 & 87.71 (00.00)  \\
                             & Outlines        & 8.05 & 8.38 & 46.57 & 12.23 & 84.64 (00.00) \\
    \cmidrule(lr){1-7}
    \textbf{Kubernetes}      & LLM only            &    NA    &       \best{0.16} &       15.32 &      0.84 &           44.38 (00.00)    \\
                             & Guidance        & \best{0.01} & 0.45 & \best{9.47} & \best{0.71} & 28.75 (04.40)  \\
                             & Llamacpp      & 0.05 & 0.28 & 28.04 & 1.06 & 28.09 (00.00)  \\
                             & Outlines        & 5.29 & 5.55 & 46.10 & 6.56 & 22.26 (00.00)  \\
    \bottomrule
    \end{tabular}
    \end{adjustbox}

    \end{table}

\begin{table}[h!]
    \centering
    \caption{\textbf{Efficiency metrics} for different engines with \textbf{Hugging Face Transformers} as the inference engine. All values are \textbf{median} of the samples.}
    \label{tab:efficiency_hf_additional}
    \renewcommand{\arraystretch}{0.9}
    \begin{adjustbox}{max width=\textwidth}
    \begin{tabular}{llrrrrr}
    \toprule
    \textbf{Dataset} & \textbf{Framework} & \textbf{GCT (s)} & \textbf{TTFT (s)} & \textbf{TPOT (ms)} & \textbf{TGT (s)} & \textbf{Output Tokens (FF)} \\
    \midrule
    \textbf{GlaiveAI}      
                            & Guidance      & \best{0.01}   & 0.36          & \best{36.92}  & \best{1.87}   & 41.45(16.76) \\
                            & XGrammar      & 0.12          & \best{0.30}   & 66.78         & 2.87          & 39.47(00.00) \\
                            
    \cmidrule(lr){1-7}
    \textbf{GitHub Easy}   
                            & Guidance      & \best{0.01}   & 0.37          & \best{42.03}  & \best{1.60}   & 27.67(06.75) \\
                            & XGrammar      & 0.11          & \best{0.33}   & 65.57         & 4.07          & 59.45(00.00) \\
                            
    \cmidrule(lr){1-7}
    \textbf{GitHub Medium}  
                            & Guidance      & \best{0.01}   & 0.55          & \best{44.21}  & \best{4.84}   & 96.31(26.93) \\
                            & XGrammar      & 0.20          & \best{0.48}   & 65.51         & 6.53          & 92.93(00.00) \\
    \cmidrule(lr){1-7}
    \textbf{GitHub Hard}    
                            & Guidance      & \best{0.01}   & 0.73          & \best{35.88}  & \best{10.25}  & 211.40(101.40) \\
                            & XGrammar      & 0.30          & \best{0.65}   & 65.20         & 14.99         & 221.40(00.00) \\
    \bottomrule
    \end{tabular}
    \end{adjustbox}

    \end{table}

\section{Quality Experiment Details}\label{app:task_experiment_details}

\paragraph{Prompt and JSON Schema}
For the task of \textbf{Shuffle Objects}, and \textbf{GSM8K}, we use the same prompt and JSON schema from the dottxt's "let me speak freely" \hyperlink{https://github.com/dottxt-ai/demos/blob/main/say-what-you-mean/GSM8K_JSON.ipynb}{rebuttal}.

For the task of \textbf{Last Letter}, we make a slight modification because the original prompt used was a bad example as pointed out by \cite{kurt2023say}. We also put it into a JSON format to better align with the other tasks.

\begin{figure}[H]
    \begin{tcolorbox}[
        colback=gray!5!white,
        colframe=gray!75!black,
        title=Prompt Template for GSM8K,
        width=\textwidth,
        boxrule=0.75pt,
        arc=2mm,
        fonttitle=\bfseries
    ]
    \textbf{System Message:} \\
    You are an expert in solving grade school math tasks. You will be presented with a grade-school math word problem and be asked to solve it. Before answering, you should reason about the problem (using the "reasoning" field in the JSON response format described below). Always respond with JSON in the format:
    \texttt{\{"reasoning": <reasoning about the answer>, "answer": <final answer>\}}. 
    The "reasoning" field contains your logical explanation, and the "answer" field contains the final numeric result. \\

    \textbf{Demo Examples:} \\

    \verb|## Input:| \texttt{\textcolor{gray}{"[example question]"}} \\
    \verb|## Output:| \texttt{\textcolor{gray}{"reasoning": "[example reasoning]", "answer": [example answer]}} \\

    ...
    
    \end{tcolorbox}
    \caption{Prompt template for solving GSM8K with JSON responses.}
    \label{fig:math-prompt-template}
\end{figure}

\begin{figure}[h]
    \centering
    \includegraphics[width=1.0\textwidth]{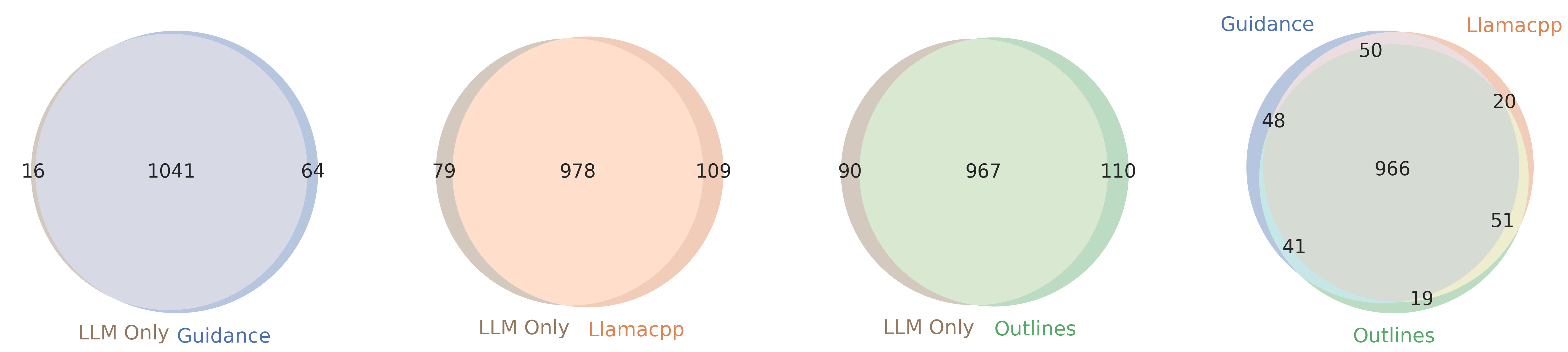}
    \caption{Overlap of Correct Instances Across Models on GSM8K}
    \label{fig:venn_diagram}
\end{figure}

Figure~\ref{fig:venn_diagram} reveals non-empty exclusive regions for each engine, indicating that no single engine outperforms the others across all instances.

\section{Engine calling Snippet}\label{app:engine_snippet}

We provide a snippet of the engine code used in our experiments. 
The \texttt{generation} method of each engine has two main components: ``\texttt{compile\_grammar}'' and ``\texttt{call\_engine}''.

\renewcommand\listoflistingscaption{List of source codes}
\begin{listing}[H]
\begin{minted}{python}
import time 
import stopit
class BaseModel:

    @stopit.threading_timeoutable(timeout=40)
    def compile_grammar(self, json_schema):
        status = "unknown"
        try:
            compiled_grammar = self._compile_grammar(json_schema)
            status = "success"
        except Exception as e:
            # Any exception in this block will be caught and considered as schema not supported
            compiled_grammar = None
            status = "schema_not_supported"
        return compiled_grammar, status

    def generate(self, prompt, json_schema=None):
        compile_start_time = time.time()
        compiled_grammar = self.compile_grammar(json_schema)
        compile_end_time = time.time()
        # GCT (Grammar Compilation Time)
        gct = compile_end_time - compile_start_time

        gen_start_time = time.time()
        output, first_tok_arr_time = self._call_engine(prompt, compiled_grammar)
        # TTFT (Time to First Token)
        ttft = first_tok_arr_time - gen_start_time
        gen_end_time = time.time()
        # TGT (Total Generation Time)
        tgt = gen_end_time - gen_start_time
        return output, gct, ttft, tgt
    
    def _call_engine(self, prompt, compiled_grammar):
        raise NotImplementedError
\end{minted}
    \caption{Abstract BaseModel interface defining the calling of structured generation, including grammar compilation and text generation timing metrics.}
    \label{listing:base_model}
\end{listing}

We use the Listing~\ref{listing:validate} to validate the generated JSONs against the schema. 
The validation is done by the \texttt{jsonschema} library with format checking enabled.

\begin{listing}[!h]
\begin{minted}{python}
import jsonschema
from jsonschema import Draft202012Validator, FormatChecker, ValidationError

format_checker = FormatChecker()

def is_json_schema_valid(schema: dict):
    try:
        jsonschema.Draft202012Validator.check_schema(schema)
        return True
    except jsonschema.SchemaError as e:
        return False

def validate_json_against_schema(json_obj, json_schema):
    if not is_json_schema_valid(json_schema):
        raise ValidationError("The JSON schema is invalid.")
    validator = Draft202012Validator(json_schema, format_checker=format_checker)
    return validator.validate(json_obj)
\end{minted}
    \caption{Validation of the generated JSONs against the schema.}
    \label{listing:validate}
\end{listing}

We provide a snippet of how the engines are called in our experiments in Listings~\ref{listing:guidance},~\ref{listing:llama_cpp},~\ref{listing:outlines}, and~\ref{listing:xgrammar}.

\begin{listing}[!h]
\begin{minted}{python}
import guidance 

class GuidanceModel(BaseModel):
    def compile_grammar(self, json_schema):
        return  guidance.json(
                schema=json_schema,
            )
    def _call_engine(self, prompt, compiled_grammar):
        generator =  self.guidance_model.stream() + prompt + compiled_grammar
        for i, state in enumerate(generator):
            if i == 0:
                first_state_arr_time = time.time()
        output = state
        return output, first_state_arr_time
\end{minted}
    \caption{Invocation of the guidance engine.}
    \label{listing:guidance}
\end{listing}

\begin{listing}[!h]
\begin{minted}{python}
import llama_cpp

class LlamaCppModel(BaseModel):

    def compile_grammar(self, json_schema):
        return  llama_cpp.llama_grammar.LlamaGrammar.from_json_schema(json_schema)
    
    def _call_engine(self, prompt, compiled_grammar):
        generator = self.llama_cpp_model.create_chat_completion(prompt, grammar=compiled_grammar, stream=True)
        output = ""
        for i, token in enumerate(generator):
            if i == 0:
                first_tok_arr_time = time.time()
            output += token
        return output, first_tok_arr_time
\end{minted}
    \caption{Invocation of the LlamaCpp engine.}
    \label{listing:llama_cpp}
\end{listing}

\begin{listing}[!h]
\begin{minted}{python}
import outlines

class OutlinesModel(BaseModel):
    def compile_grammar(self, json_schema):
        return  outlines.generate.json(
                schema_object=json_schema
            )
    def _call_engine(self, prompt, compiled_grammar):
        generator =  self.generator.stream(prompt)
        output = ""
        for i, token in enumerate(generator):
            if i == 0:
                first_tok_arr_time = time.time()
            output += token
        return output, first_tok_arr_time
               
\end{minted}
    \caption{Invocation of the Outlines engine.}
    \label{listing:outlines}
\end{listing}

\begin{listing}[!h]
\begin{minted}{python}
import xgrammar

class TimingLogitsProcessor(LogitsProcessor):
    def __init__(self):
        super().__init__()
        self.timestamps = []

    def __call__(self, input_ids, scores):
        current_time = time.time()
        self.timestamps.append(current_time)
        return scores
    
class XGrammarModel(BaseModel):
    def compile_grammar(self, json_schema):
        return  xgrammar.GrammarCompiler().compile_json_schema(json_schema)
    
    def _call_engine(self, prompt, compiled_grammar):
        output = self.hf_model.generate(prompt, logits_processor=[compiled_grammar, timeit_logit_processor])
        first_tok_arr_time = timeit_logit_processor.timestamps[0]
        return output, first_tok_arr_time
\end{minted}
    \caption{Invocation of the XGrammar engine.}
    \label{listing:xgrammar}
\end{listing}

\end{document}